\documentclass{article} 
\usepackage[final]{colm2025_conference}

\usepackage[inkscapelatex=false]{svg}

\usepackage{color,xcolor}
\usepackage{epsfig}
\usepackage{graphicx}
\usepackage{graphics}
\usepackage{inconsolata}
\usepackage{lineno}
\usepackage{xstring}

\usepackage{hyperref}
\usepackage{url}
\hypersetup{pagebackref=true,breaklinks=true,letterpaper=true,colorlinks, citecolor=darkblue,bookmarks=false}

\usepackage{adjustbox}
\usepackage{array}
\usepackage{booktabs}
\usepackage{colortbl}
\usepackage{float,wrapfig}
\usepackage{hhline}
\usepackage{multirow}
\usepackage{subcaption} 
\usepackage[toc,page]{appendix}
\usepackage{tabularx}
\usepackage{tcolorbox}

\usepackage{bbm}
\usepackage{amsmath,amsfonts,amsthm,amssymb}
\usepackage{bm}
\usepackage{nicefrac}
\usepackage{microtype}
\usepackage{inconsolata}
\usepackage{pifont}
\usepackage{dsfont}
\usepackage{fontawesome5}

\usepackage{changepage}
\usepackage{extramarks}
\usepackage{fancyhdr}
\usepackage{lastpage}
\usepackage{setspace}
\usepackage{soul}
\usepackage{xspace}
\usepackage{indentfirst}
\usepackage{paralist}

\usepackage{algorithm}
\usepackage{enumerate}
\usepackage{lipsum}
\usepackage{physics}
\usepackage{cleveref}

\usepackage{listings}
\usepackage{xcolor}
\usepackage{enumitem}
\usepackage{fancyvrb}
\usepackage{fvextra}
\usepackage{algpseudocode}
\usepackage{placeins}

\definecolor{commentgray}{rgb}{0.5, 0.5, 0.5}

\newcolumntype{L}[1]{>{\raggedright\let\newline\\\arraybackslash\hspace{0pt}}m{#1}}
\newcolumntype{C}[1]{>{\centering\let\newline\\\arraybackslash\hspace{0pt}}m{#1}}
\newcolumntype{R}[1]{>{\raggedleft\let\newline\\\arraybackslash\hspace{0pt}}m{#1}}

\newcommand{\ignorethis}[1]{}

\makeatletter
\DeclareRobustCommand\onedot{\futurelet\@let@token\@onedot}
\def\@onedot{\ifx\@let@token.\else.\null\fi\xspace}

\def\eg{\emph{e.g}\onedot}

\makeatother

\makeatletter
\def\adl@drawiv#1#2#3{%
        \hskip.5\tabcolsep
        \xleaders#3{#2.5\@tempdimb #1{1}#2.5\@tempdimb}%
                #2\z@ plus1fil minus1fil\relax
        \hskip.5\tabcolsep}
\newcommand{\cdashlinelr}[1]{%
  \noalign{\vskip\aboverulesep
           \global\let\@dashdrawstore\adl@draw
           \global\let\adl@draw\adl@drawiv}
  \cdashline{#1}
  \noalign{\global\let\adl@draw\@dashdrawstore
           \vskip\belowrulesep}}
\makeatother

\definecolor{highlight}{HTML}{71b7ed}
\definecolor{second}{HTML}{B8DBF6}
\definecolor{customBlue}{RGB}{113,183,237}
\definecolor{customGreen}{RGB}{113,237,183}

\theoremstyle{plain}

\theoremstyle{definition}

\theoremstyle{remark}

\newcommand{\benchmarkname}{\text{CrossWordBench}}
\newcommand{\err}[2]{#1\ensuremath{_{\pm #2}}}

\definecolor{citecolor}{HTML}{0071bc}
\definecolor{mydarkblue}{rgb}{0,0.08,1}
\definecolor{mydarkgreen}{rgb}{0.02,0.6,0.02}
\definecolor{mydarkred}{rgb}{0.8,0.02,0.02}
\definecolor{mydarkorange}{rgb}{0.40,0.2,0.02}
\definecolor{mypurple}{RGB}{111,0,255}
\definecolor{myred}{rgb}{1.0,0.0,0.0}
\definecolor{mygold}{rgb}{0.75,0.6,0.12}
\definecolor{mydarkgray}{rgb}{0.66, 0.66, 0.66}

\definecolor{darkblue}{rgb}{0, 0, 0.5}
\definecolor{darkgreen}{rgb}{0.02,0.6,0.02}
\definecolor{darkred}{rgb}{0.8,0.02,0.02}
\definecolor{darkorange}{rgb}{0.40,0.2,0.02}
\definecolor{darkpurple}{RGB}{111,0,255}
\definecolor{linkcolor}{RGB}{200, 0, 130}

\newif\ifarxiv
\arxivtrue

\ifarxiv
  \renewcommand{\includesvg}[2][]{%
    \StrSubstitute{#2}{graphs/}{pdf_graphs/}[\tempfile]%
    \StrSubstitute{\tempfile}{.svg}{.pdf}[\finalfile]%
    \includegraphics[#1,trim=0 0 0 0,clip]{\finalfile}%
  }
\fi

\title{\raisebox{-0.3\height}{\includegraphics[height=1cm]{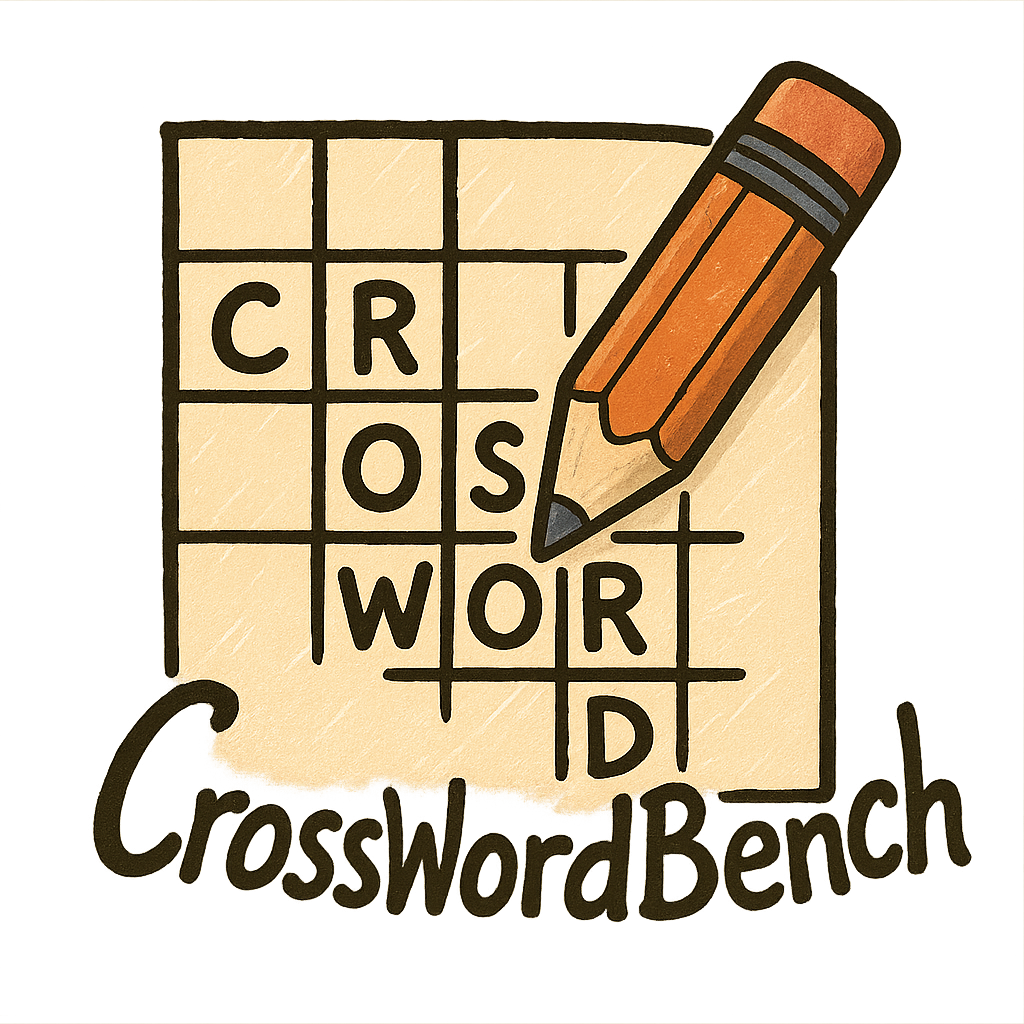}}CrossWordBench: Evaluating the Reasoning Capabilities of LLMs and LVLMs with Controllable Puzzle Generation}

\author{
\vspace{-2.5em}\\
\textbf{Jixuan Leng$^{1}$, Chengsong Huang$^2$, Langlin Huang$^2$, Bill Yuchen Lin$^3$,} \\
\textbf{William W. Cohen$^1$, Haohan Wang$^4$, Jiaxin Huang$^2$} \\
$^1$CMU, $^2$WUSTL, $^3$UW, $^4$UIUC \\
\texttt{jixuanl@cs.cmu.edu, jiaxinh@wustl.edu} \\\\
\vspace{-1.5em} \\
\centerline{\textbf{Code}: {\url{https://github.com/SeanLeng1/CrossWordBench}}}\\
\centerline{\textbf{Dataset}: {\url{https://huggingface.co/datasets/HINT-lab/CrossWordBench}}}
}

\begin{document}

\ifcolmsubmission
\linenumbers
\fi

\maketitle

\vspace{-1.5em}
\begin{abstract}
Existing reasoning evaluation frameworks for Large Language Models (LLMs) and Large Vision-Language Models (LVLMs) predominantly assess either text-based reasoning or vision-language understanding capabilities, with limited dynamic interplay between textual and visual constraints.
To address this limitation, we introduce \benchmarkname{}, a benchmark designed to evaluate the reasoning capabilities of both LLMs and LVLMs through the medium of crossword puzzles---a task requiring multimodal adherence to semantic constraints from \textbf{text-based clues} and intersectional constraints from \textbf{visual grid structures}.
\benchmarkname{} leverages a controllable puzzle generation framework that produces puzzles in two formats (\textit{text} and \textit{image}), supports adjustable difficulty through prefill ratio control, and offers different evaluation strategies, ranging from direct puzzle solving to interactive modes suitable for agentic evaluation.
Our extensive evaluation of over 20 models reveals that reasoning LLMs substantially outperform non-reasoning models by effectively leveraging crossing-letter constraints.
We further demonstrate that LVLMs struggle with the task, showing a strong correlation between their puzzle-solving performance and grid-parsing accuracy. 
Our findings highlight the limitations of the reasoning capabilities of current LLMs and LVLMs, and provide an effective approach for creating multimodal constrained tasks for future evaluations.


\end{abstract}

\section{Introduction}
\begin{figure}[htbp]
    \centering
    \vspace{-1.0em}
    \captionsetup{aboveskip=-1.8pt, belowskip=-1.8pt}
    \includesvg[width=1.0\textwidth]{graphs/teaser.svg}
    \caption{Performance gap between top-performing reasoning LLMs and non-reasoning LLMs/LVLMs on CrossWordBench. Reasoning LLMs achieve better overall performance and adhere more effectively to crossing-letter constraints than non-reasoning LLMs/LVLMs.
    }
    \label{fig:intro}
    \vspace{-0.5em}
\end{figure}

Large reasoning models (\eg, OpenAI-o1~\citep{jaech2024openai} and DeepSeek-R1~\citep{guo2025deepseek}) have made exceptional progress on reasoning benchmarks including math problem solving~\citep{aime_1983_2024}, coding~\citep{Austin2021ProgramSW} and commonsense reasoning~\citep{allenai:arc}. However, existing evaluation benchmarks for Large Language Models (LLMs) and Large Vision-Language Models (LVLMs) (\eg, Visual Question Answering~\citep{Yue2023MMMUAM}) mostly focus on text-based reasoning or vision-language understanding, lacking dynamic interplay between textual and visual constraints that characterizes real-world problem solving. Consequently, evaluating multimodal reasoning capabilities---particularly on reasoning tasks requiring both textual and visual constraints---remains challenging. 

\textbf{Crossword puzzles}, a classic grid-based task in which horizontal (``Across'') words and vertical (``Down'') words must be filled in based on text-based clues, provide a unique testbed for such evaluations. They pose two distinct challenges: (1) question answering for each \textbf{text-based clue}, which may admit multiple correct solutions, and (2) \textbf{visual constraint satisfaction}, which requires precise letter alignment at the intersections of Across and Down entries. Prior crossword datasets~\citep{efrat2021cryptonite, rozner2021decrypting, kulshreshtha2022down, chen2025lr} often suffer from their reliance on static and copyrighted online news sources and adopt a text-centric formulation, thereby neglecting the visual structure.

In this paper, we introduce \benchmarkname{}, a controllable and scalable benchmark for evaluating the reasoning capabilities of both LLMs and LVLMs. \benchmarkname{} collects data and generates puzzles from three sources: (1) multilingual word-clue pairs from public repositories, (2) dictionary-based definitions, and (3) adapted question-answer pairs from existing benchmarks (\eg, CommonsenseQA~\citep{talmor2018commonsenseqa}) where the answers are open-ended or unconstrained. 
By representing grids and clues in different formats, as shown in Figure~\ref{fig: data-pipeline}, \benchmarkname{} facilitates the evaluation of both model types.
It supports two evaluation modes: a direct puzzle-solving mode, which generates one-round responses using zero-shot Chain-of-Thought (CoT) prompts, and an interactive mode for step-by-step puzzle-solving, where grid update functions can provide intermediate visual outputs for follow-ups, thereby serving as a foundation for evaluating agents using function calling.

We evaluate over 20 state-of-the-art models, including both proprietary models (\eg, GPT-4o~\citep{hurst2024gpt} and Claude 3.7 Sonnet~\citep{anthropic_claude_2024}) and open-weight models (\eg, DeepSeek-R1~\citep{guo2025deepseek} and Pixtral-Large-Instruct~\citep{agrawal2024pixtral}). 
Our evaluation yields several notable findings: (1) \textbf{LVLMs perform significantly worse than LLMs} on \benchmarkname{} (as shown in Figure~\ref{fig:intro}), and they struggle in OCR in vertical (``Down'') word extraction. In fact, their puzzle-solving performance strongly correlates with their grid-parsing accuracy ($r = 0.94$). (2) Reasoning LLMs outperform non-reasoning models, and \textbf{benefit from both test-time scaling and increased crossing-letter constraints}. (3) Even \textbf{puzzles derived from saturated benchmarks} (\eg, CommonsenseQA) \textbf{remain challenging}, highlighting the significance of structural constraints in reasoning evaluation.

\section{Related Work}
\vspace{-0.2em}

\textbf{LLMs and LVLMs Reasoning.}
Recent research on the reasoning capabilities of LLMs~\citep{ahn2024large, huang2022towards, kojima2022large, plaat2024reasoning, jaech2024openai, guo2025deepseek, o3_mini, huang2022large} has led to the development of various approaches, including prompting-based methods~\citep{wei2022chain, yao2023tree, besta2024graph, chen2022program} that guide LLMs through intermediate reasoning steps for solving complex problems, fine-tuning methods that train LLMs on long reasoning chains~\citep{ye2025limo, muennighoff2025s1, zhao202514millionopensourcedistilled}, and test-time scaling via self-refinement or the use of a verifier~\citep{setlur2024rewarding, feng2023alphazero, wang2023math, zhang2024rest, uesato2022solving, huang2025efficient}.
Moreover, a recent study from Deepseek-R1~\citep{guo2025deepseek} demonstrates that reinforcement learning (RL) with verifiable rewards facilitates the emergence of complex thinking processes in LLMs. 
Several studies have explored reasoning in the multimodal domain by constructing CoT data for fine-tuning~\citep{xu2024llava} and developing verifiable problems for RL~\citep{yang2025r1, huang2025vision}.
While these approaches have demonstrated success, existing evaluation datasets remain largely restricted to math problems~\citep{wang2024measuring, lu2023mathvista, zhang2024mathverse}.

\textbf{Crossword Puzzles in Language Model Evaluation.}
Crossword puzzles have long been a focus of research in natural language processing (NLP), particularly before the advent of LLMs.
Early approaches typically employed constraint satisfaction algorithms augmented by external knowledge bases.
Notable examples include systems such as \textit{Proverb}~\citep{littman2002probabilistic}, \textit{Dr. Fill}~\citep{ginsberg2011dr}, and specialized models such as \textit{the Berkeley Crossword Solver}~\citep{wallace2022automated}, which incorporate a fine-tuned BERT~\citep{devlin2019bert} and belief propagation.
More recent studies have leveraged LLMs to address crossword puzzles through techniques including fine-tuning~\citep{efrat2021cryptonite, rozner2021decrypting, sadallah2024llms}, prompting strategies such as Tree-of-Thoughts~\citep{yao2023tree}, or integration with search algorithms~\citep{saha2024language}, demonstrating the potential of LLMs for crosswords.

Several datasets for crossword puzzles have been proposed, covering both English~\citep{efrat2021cryptonite, rozner2021decrypting, kulshreshtha2022down, chen2025lr} and Arabic~\citep{zeinalipour2025arabic}.
However, one significant limitation of these datasets and approaches is that they rely on data from online news sources~\citep{efrat2021cryptonite, rozner2021decrypting, kulshreshtha2022down, chen2025lr} and often formulate crossword solving as a question-answering (QA) task~\citep{sadallah2024llms, efrat2021cryptonite, rozner2021decrypting, yao2023tree}, thereby overlooking the fundamental constraint-based nature of the problem. Moreover, all of them treat crossword solving as a text-based task, despite the inherently visual nature of crossword grids, leaving a gap in extending crossword puzzles for evaluating LVLMs.
In this work, we address these limitations by introducing CrossWordBench, a framework that features controllable puzzle generation and extends evaluation to LVLMs.

\section{Benchmark Curation}\label{data curation}
\begin{figure}[htbp]
    \centering
    \vspace{-1.0em}
    \includesvg[width=1.0\textwidth]{graphs/data-pipeline.svg}
    \caption{Framework of CrossWordBench. (a) Dataset curation process and input templates for LLMs and LVLMs; (b) Zero-shot CoT evaluation; (c) Interactive Mode Evaluation.}
    \label{fig: data-pipeline}
    \vspace{-0.7em}
\end{figure}
Online sources, including \textit{The New York Times} and \textit{The Los Angeles Times}, provide complete crossword puzzles.
Nevertheless, directly utilizing these puzzles presents several disadvantages.
\textbf{(1)} Copyright restrictions may limit their usage, and their online availability increases the probability that they have been incorporated into the pretraining datasets of contemporary models.
\textbf{(2)} Online puzzles are typically static---with predefined words, themes, grid sizes, and strict formatting (e.g., black square symmetry)\smash{\footnotemark[1]}---which not only limits their adaptability for diverse benchmark generation, but also imposes arbitrary formatting constraints that do not yield meaningful benefits for evaluation.
Prior research~\citep{mirzadeh2024gsm} has shown that even minor modifications to questions can significantly affect model performance, particularly when performance on the original version is near saturation.
\textbf{(3)} The static characteristics of these puzzles restrict the range of potential evaluation strategies.
To overcome these limitations, we propose a two-stage benchmark construction strategy:

\footnotetext[1]{\footnotesize \href{https://www.nytimes.com/article/submit-crossword-puzzles-the-new-york-times.html\#link-18b4d122}{The New York Times Crossword Requirements.}}
\footnotetext[2]{\footnotesize \href{https://cryptics.georgeho.org/}{A Dataset of Cryptic Crossword Clues}}
\footnotetext[3]{\footnotesize \href{https://github.com/yaozheng/crossword}{Chinese Crosswords}}

\textbf{Word-Clue Pairs Curation.}
We compile word-clue pairs from three source categories:
\textbf{(1) Public repositories:} We selectively extract individual, often cryptic, word-clue pairs from public online repositories with samples in both English\smash{\footnotemark[2]} and Chinese\smash{\footnotemark[3]}.
\textbf{(2) Dictionary-Based Pairs:} 
We collect standard English words and use their dictionary definitions from NLTK WordNet\smash{\footnotemark[4]} as clues, referring to this category as \textbf{English Simple}.
\textbf{(3) Adapted Benchmark Data:} We demonstrate that existing LLM benchmarks, which are typically designed for open-ended or multiple-choice questions without strict formatting constraints, can be transformed into word-clue pairs. In our study, we filter the CommonsenseQA~\citep{talmor2018commonsenseqa} \textbf{training} set for single-word answers and use the associated questions as clues.

\footnotetext[4]{\footnotesize \href{https://www.nltk.org/howto/wordnet.html}{NLTK Wordnet}}

While the possibility of data contamination may still exist for these word-clue pairs, the variations in grid design can still yield distinct crossword puzzles, and as demonstrated in Section~\ref{experiment_extension}, even when incorporating extensively tuned data such as CommonsenseQA training set, the resulting puzzles remain remarkably challenging for both LLMs and LVLMs.
\begin{table}[htbp]
\footnotesize
  \centering
  \vspace{-1.5em}
  \caption{Crossword puzzle statistics for different subjects and grid sizes. Statistics are presented separately for each category, as distinct word and clue pairs are used for their construction. Additionally, the aggregated statistics for all English puzzles are included.}
      \begin{tabularx}{\textwidth}{@{}Xccccc}
        \toprule
        \multirow{2}{*}{\textbf{Stats.}} & \multicolumn{2}{c}{\textbf{English}} & \textbf{Chinese} & \textbf{English Simple} & \textbf{CommonsenseQA} \\
        \cmidrule(lr){2-3}
         & 7×7 & 14×14 & 7×7 & 7×7 & 7×7 \\
        \midrule
        Total $\#$ of puzzles               & 100       & 100       & 100       & 100       & 50        \\
        Total $\#$ of words                 & 1,193     & 3,472      & 1,327      & 1,139      & 543       \\
        Unique words (\% total)     & 83.82\%   & 80.76\%   & 92.92\%   & 36.70\%   & 59.85\%   \\
        Unique clues (\% total)     & 100\%     & 100\%     & 100\%     & 100\%     & 100\%     \\
        \midrule
        \multicolumn{6}{l}{\footnotesize\textbf{Aggregated (English 7x7 \& 14x14): 200 puzzles, 74.02\% unique words, 100\% unique clues}}\\
        \midrule
        \multicolumn{6}{l}{\textit{Words per Puzzle}} \\
        \quad Minimum               & 11        & 22        & 11        & 11        & 9         \\
        \quad Maximum               & 16        & 44        & 18        & 13        & 13        \\
        \quad Mean                  & 11.93     & 34.72     & 13.27     & 11.39     & 10.86     \\
        \midrule
        \multicolumn{6}{l}{\textit{Word Length (Letters)}} \\
        \quad Minimum               & 2         & 3         & 2         & 3         & 3         \\
        \quad Maximum               & 5         & 12        & 5         & 5         & 5         \\
        \quad Mean                  & 3.59      & 4.31      & 3.02      & 3.63      & 3.77      \\
        \midrule
        Avg blocked cells (\%)      & 39.51\%   & 45.22\%   & 43.37\%   & 39.12\%   & 38.37\%   \\
        \bottomrule
      \end{tabularx}
      \vspace{-1.5em}
      \label{tab:crossword_stats_all}
\end{table}

\textbf{Puzzle Generation.}
After collecting word-clue pairs, we implement an automatic pipeline to generate puzzles in both \textbf{text} and \textbf{image} formats, as illustrated in Figure~\ref{fig: data-pipeline}. The pipeline enables (1) puzzle difficulty control by adjusting grid sizes, and (2) incremental word placement functions, which facilitate the interactive evaluation mode described in Section~\ref{experiment: interactive mode}.

The generation framework leverages a heuristic scoring function that encourages desirable properties such as a high number of horizontal and vertical word intersections. Post-processing filters are applied to discard puzzles with low clue counts or high blocked cell ratios, thereby enhancing overall quality and structural coherence. These heuristics and filters, in combination with grid size adjustments, help control puzzle complexity.
To preserve the category integrity, we ensure that clues do not overlap within each category\smash{\footnotemark[5]}.

\footnotetext[5]{\footnotesize Please check out our \href{https://github.com/SeanLeng1/CrossWordBench}{code} for more implementation details.}

\textbf{Puzzle Quality.}
While CrossWordBench does not include human-created puzzles, its design is inspired by them. Table~\ref{tab:crossword_stats_all} reports overall dataset statistics---including clue count, average word length, and blocked cell ratio---which serve as proxies for puzzle quality and difficulty.

Compared human-created puzzles---for example, those used LR$^{2}$Bench~\citep{chen2025lr}, which were sourced from outlets such as \textit{The Los Angeles Times} and \textit{Vulture}---our generated puzzles exhibit comparable difficulty when evaluated using similar word-level metrics.
Our focus on automatically generated puzzles is motivated by the need for scalable, controllable, and diverse evaluation across subjects and formats. This enables systematic variation in difficulty through grid size, prefill ratio, and clue selection; supports step-by-step reasoning via an interaction mode; and facilitates fine-grained metric analysis beyond overall accuracy.

\vspace{-0.5em}
\section{Experiment}\label{Experiment}
\vspace{-0.5em}
In the following sections, we present an extensive empirical evaluation of \benchmarkname{} across multiple different model architectures and analyze their performance characteristics.
 
\subsection{Experimental Setup}

\textbf{Evaluation Data and Metrics.}
For the main experiments, we evaluate models on the \textbf{English} set
with three metrics to assess answer accuracy and adherence to crossword constraints.
\begin{itemize}[leftmargin=*, itemsep=0pt, topsep=0pt]
\item \textbf{Word Coverage Rate (WCR)}: word-level accuracy, percentage of correctly solved words.                 

\item \textbf{Letter Coverage Rate (LCR)}: letter-level accuracy, percentage of correct letter placements.

\item \textbf{Intersection Consistency Rate (ICR)}:
the internal consistency of the model's answers at intersections where across and down words overlap, defined as:
\vspace*{-0.2\baselineskip}
\begin{equation}
\mathrm{ICR}=\frac{1}{|\mathcal{I}|} \sum_{(a, d, j, k) \in \mathcal{I}} \mathbb{1}\{a[j]=d[k]\}
\end{equation}\label{ICR}where $\mathcal{I}$ denotes the set of all intersections, where each tuple $(a, d, j, k)$ indicates the $j$th letter of the across word $a$ overlaps with the $k$th letter of the down word $d$.
This metric reflects whether models correctly adhere to the grid structural constraints of a puzzle.
\end{itemize}

\textbf{Evaluated Models.} 
We evaluate a collection of proprietary and open-weight models, including both LVLMs and LLMs.
For \textbf{proprietary models}, we consider state-of-the-art models such as GPT-4o~\citep{hurst2024gpt} and Claude 3.7 Sonnet~\citep{anthropic_claude_2024} (with and without thinking mode). 
For \textbf{open-weight models}, our selections range from 3B to 124B parameters, such as the {Qwen} Series~\citep{qvq-72b-preview, bai2025qwen2} and {Pixtral-Large-Instruct-2411}~\citep{agrawal2024pixtral} for LVLMs. 
For LLMs, we include both reasoning models such as the {Deepseek} Series~\citep{liu2024deepseek, guo2025deepseek} and non-reasoning models such as {Llama} series~\citep{dubey2024llama}. The full list of models is shown in Table~\ref{tab:main_results}.
We set decoding temperature to 0 for non-reasoning models for consistency, 0.6 for reasoning models based on commonly recommended settings, and 1.0 for certain proprietary models (e.g., used by Claude 3.7-Sonnet with Thinking). Further generation details are provided in Appendix~\ref{appendix_generation_details}.

\textbf{Input Prompt Templates and Output Response Parsing.}
For the main evaluation, we adopt the zero-shot Chain-of-Thought (CoT)~\citep{wei2022chain} prompting strategy. In LVLM evaluation, 
the clues and the grid are both embedded within an image. In LLM evaluation, the grid is represented as a 2D binary array, with \texttt{1} indicating a blocked cell and \texttt{0} representing an unfilled cell, and is prepended to text clues in the prompt.
To extract answers from responses, we leverage the structured output capabilities of o3-mini\smash{\footnotemark[6]} to convert raw model responses into JSON format by generating dynamic Pydantic models. Detailed prompt templates and implementation details are listed in Appendix~\ref{appendix prompts} and~\ref{appendix parsing}.

\footnotetext[6]{\footnotesize \url{https://openai.com/index/openai-o3-mini/}}



\subsection{Main Results}
\begin{table}[htbp]
    \centering
    \footnotesize
    \caption{Comparison of various LLMs and LVLMs on CrossWordBench \textbf{English} set across two difficulty levels using zero-shot CoT. We report the mean and standard error over 100 samples for both \texttt{7x7} and \texttt{14x14} grids. \protect\smash{\faLightbulb[regular]} indicates that the model is a reasoning model. \protect\smash{$^{\dagger}$}: We use the \href{https://fireworks.ai/}{Fireworks API} for DeepSeek V3 and Llama-3.1-405B, while \href{https://api-docs.deepseek.com/}{offical API} for R1.}
    \setlength{\tabcolsep}{2pt} 
    \renewcommand{\arraystretch}{1.3} 
    \resizebox{\linewidth}{!}{
    \begin{tabularx}{\textwidth}{@{}Xccc|ccc}
        \toprule
        \multirow{2}{*}{\textbf{Models}} & \multicolumn{3}{c}{\textbf{7x7}} & \multicolumn{3}{c}{\textbf{14x14}} \\
        \cmidrule(lr){2-4} \cmidrule(lr){5-7}
         & \textbf{WCR} & \textbf{LCR} & \textbf{ICR} 
         & \textbf{WCR} & \textbf{LCR} & \textbf{ICR}  \\
        \midrule[\heavyrulewidth]
        \addlinespace[0.5em]
        \multicolumn{7}{c}{\cellcolor{customBlue}\textit{\textbf{Proprietary LVLMs}}} \\
        \midrule
        
        \bf\mbox{Claude-3-7-Sonnet}                                                 & \err{0.479}{0.014} & \err{0.528}{0.013} & \err{0.366}{0.016} & \err{0.416}{0.009} & \err{0.449}{0.009} & \err{0.272}{0.010} \\
        
        \bf\mbox{Claude-3-7-Sonnet \faLightbulb[regular]}                           & \err{0.365}{0.017} & \err{0.448}{0.014} & \err{0.330}{0.015} & \err{0.382}{0.009} & \err{0.428}{0.007} & \err{0.228}{0.008} \\
        
        \bf\mbox{GPT-4o-2024-11-20}                                                 & \err{0.348}{0.015} & \err{0.403}{0.015} & \err{0.234}{0.017} & \err{0.350}{0.009} & \err{0.393}{0.008} & \err{0.190}{0.008} \\
        
        \bf\mbox{Gemini 2.0 Pro Exp}                                                & \err{0.351}{0.014} & \err{0.368}{0.014} & \err{0.339}{0.015} & \err{0.273}{0.008} & \err{0.303}{0.007} & \err{0.221}{0.007} \\
        
        \bf\mbox{Gemini 2.0 Flash}                                                  & \err{0.277}{0.015} & \err{0.300}{0.013} & \err{0.225}{0.013} & \err{0.260}{0.008} & \err{0.284}{0.008} & \err{0.190}{0.007} \\
        \midrule[\heavyrulewidth]
        \addlinespace[0.5em]
        \multicolumn{7}{c}{\cellcolor{customBlue}\textit{\textbf{Open-Weight LVLMs}}} \\
        \midrule
        \bf\mbox{Pixtral-Large-Instruct-2411}                                   & \err{0.297}{0.015} & \err{0.338}{0.014} & \err{0.198}{0.014} & \err{0.251}{0.009} & \err{0.284}{0.007} & \err{0.134}{0.007} \\
        
        \bf\mbox{InternVL2\_5-78B-MPO}                                          & \err{0.121}{0.011} & \err{0.164}{0.009} & \err{0.099}{0.011} & \err{0.119}{0.007} & \err{0.159}{0.006} & \err{0.073}{0.005} \\

        \bf\mbox{NVLM-D-72B}                                                    & \err{0.134}{0.010} & \err{0.179}{0.009} & \err{0.076}{0.008} & \err{0.085}{0.006} & \err{0.120}{0.007} & \err{0.053}{0.004} \\
        
        \bf\mbox{Qwen2.5-VL-72B-Instruct}                                       & \err{0.207}{0.013} & \err{0.245}{0.011} & \err{0.133}{0.011} & \err{0.194}{0.007} & \err{0.227}{0.006} & \err{0.110}{0.006} \\
        
        \bf\mbox{QVQ-72B-Preview}                                               & \err{0.197}{0.012} & \err{0.218}{0.010} & \err{0.091}{0.008} & \err{0.195}{0.007} & \err{0.215}{0.007} & \err{0.108}{0.006} \\
        
        \bf\mbox{llava-onevision-72b-ov-chat}                                   & \err{0.141}{0.012} & \err{0.165}{0.010} & \err{0.097}{0.009} & \err{0.112}{0.008} & \err{0.141}{0.007} & \err{0.075}{0.005} \\

        \bf\mbox{gemma-3-27b-it}                                                & \err{0.158}{0.011} & \err{0.218}{0.011} & \err{0.124}{0.010} & \err{0.106}{0.009} & \err{0.160}{0.009} & \err{0.075}{0.005} \\
        
        \bf\mbox{Aria}                                                          & \err{0.061}{0.009} & \err{0.101}{0.007} & \err{0.051}{0.006} & \err{0.035}{0.006} & \err{0.070}{0.006} & \err{0.046}{0.004} \\
        
        \bf\mbox{MiniCPM-V-2\_6}                                                & \err{0.043}{0.007} & \err{0.085}{0.006} & \err{0.064}{0.008} & \err{0.023}{0.004} & \err{0.057}{0.004} & \err{0.040}{0.003} \\
        
        \bf\mbox{Qwen2.5-VL-3B-Instruct}                                        & \err{0.013}{0.003} & \err{0.040}{0.004} & \err{0.038}{0.006} & \err{0.014}{0.002} & \err{0.034}{0.003} & \err{0.023}{0.003} \\

        \midrule[\heavyrulewidth]
        \addlinespace[0.5em]
        \multicolumn{7}{c}{\cellcolor{customGreen}\textit{\textbf{Proprietary LLMs}}} \\
        \midrule

        \bf\mbox{o3-mini-high \faLightbulb[regular]}                                & \err{0.587}{0.023} & \err{0.684}{0.021} & \err{0.891}{0.018} & \err{0.445}{0.011} & \err{0.520}{0.011} & \err{0.512}{0.007} \\

        \bf\mbox{Claude-3-7-Sonnet \faLightbulb[regular]}                           & \err{0.617}{0.019} & \err{0.712}{0.017} & \err{0.754}{0.021} & \err{0.492}{0.013} & \err{0.542}{0.012} & \err{0.431}{0.014} \\

        \bf\mbox{Claude-3-7-Sonnet}                                                 & \err{0.482}{0.015} & \err{0.574}{0.014} & \err{0.472}{0.019} & \err{0.446}{0.011} & \err{0.485}{0.011} & \err{0.321}{0.011} \\
        
        \bf\mbox{GPT-4o-2024-11-20}                                                 & \err{0.410}{0.018} & \err{0.472}{0.018} & \err{0.288}{0.019} & \err{0.338}{0.011} & \err{0.369}{0.012} & \err{0.196}{0.010} \\
        
        \bf\mbox{Gemini 2.0 Pro Exp}                                                & \err{0.460}{0.014} & \err{0.525}{0.012} & \err{0.388}{0.016} & \err{0.425}{0.009} & \err{0.457}{0.008} & \err{0.289}{0.010} \\
        
        \bf\mbox{Gemini 2.0 Flash}                                                  & \err{0.301}{0.014} & \err{0.318}{0.012} & \err{0.255}{0.014} & \err{0.280}{0.007} & \err{0.298}{0.006} & \err{0.198}{0.006} \\

        \midrule[\heavyrulewidth]
        \addlinespace[0.5em]
        \multicolumn{7}{c}{\cellcolor{customGreen}\textit{\textbf{Open-Weight LLMs}}} \\
        \midrule
        \bf\mbox{Llama-3.1-405B-Instruct}$^{\dagger}$                           & \err{0.161}{0.013} & \err{0.359}{0.012} & \err{0.243}{0.017} & \err{0.355}{0.013} & \err{0.390}{0.009} & \err{0.222}{0.008} \\
        
        \bf\mbox{DeepSeek-R1 \faLightbulb[regular]}                             & \err{0.646}{0.019} & \err{0.707}{0.017} & \err{0.678}{0.023} & \err{0.472}{0.011} & \err{0.507}{0.011} & \err{0.356}{0.011} \\
        
        \bf\mbox{DeepSeek-V3}$^{\dagger}$                                       & \err{0.303}{0.014} & \err{0.369}{0.014} & \err{0.186}{0.013} & \err{0.290}{0.009} & \err{0.335}{0.008} & \err{0.145}{0.007} \\
        
        \bf\mbox{R1-Distill-Llama-70B \faLightbulb[regular]}                     & \err{0.387}{0.015} & \err{0.448}{0.015} & \err{0.347}{0.017} & \err{0.285}{0.009} & \err{0.319}{0.009} & \err{0.161}{0.008} \\
        
        \bf\mbox{Llama-3.3-70B-Instruct}                                        & \err{0.303}{0.013} & \err{0.371}{0.012} & \err{0.206}{0.014} & \err{0.280}{0.011} & \err{0.340}{0.009} & \err{0.173}{0.008} \\
        
        \bf\mbox{QwQ-32B \faLightbulb[regular] }                                & \err{0.347}{0.017} & \err{0.445}{0.018} & \err{0.518}{0.020} & \err{0.254}{0.009} & \err{0.307}{0.009} & \err{0.189}{0.009} \\
        
        \bf\mbox{Open-Reasoner-Zero-32B \faLightbulb[regular]}                  & \err{0.139}{0.010} & \err{0.204}{0.010} & \err{0.184}{0.012} & \err{0.146}{0.007} & \err{0.199}{0.007} & \err{0.095}{0.005} \\
            
        \bf\mbox{Phi-4}                                                         & \err{0.122}{0.010} & \err{0.194}{0.010} & \err{0.113}{0.011} & \err{0.140}{0.007} & \err{0.200}{0.006} & \err{0.085}{0.005} \\
        \bottomrule
    \end{tabularx}}
    \label{tab:main_results}
\end{table}

\textbf{Reasoning LLMs substantially outperform conventional ones across metrics, with notable improvements in ICR}, as shown in Table~\ref{tab:main_results}.
In particular, among reasoning LLMs, o3-mini achieves an ICR of 0.891 on \texttt{7x7} grids, demonstrating a strong ability to interpret and enforce grid constraints. Although DeepSeek-R1 attains the highest WCR and LCR on \texttt{7x7} grids, its ICR is slightly lower than that of o3-mini.
Other reasoning models, such as QwQ-32B and R1-Distilled-Llama-70B, show moderate performance, with WCRs of approximately 0.347 and 0.387, respectively.
One notable outlier is Open-Reasoner-Zero-32B, which performs poorly across all metrics, indicating that it does not effectively leverage grid constraints for reasoning. 
This suggests that its training---primarily focused on mathematical reasoning---does not generalize well to tasks requiring spatial and linguistic integration, thereby highlighting a key limitation in the training strategies of these reasoning models. 

\textbf{Non-reasoning LLMs show limitations in ICR, and performance declines further with increasing grid size.}
Among non-reasoning LLMs, Claude 3.7 Sonnet and Gemini 2.0 Pro Exp yield the best results, with WCRs of 0.482 and 0.460 on the \texttt{7x7} grid, respectively; however, their relatively lower ICR indicates model limitations on explicit reasoning constraints for crossword puzzles. Notably, thinking mode improves Claude 3.7 Sonnet on all three metrics, highlighting the importance of reasoning and reflection in solving constraints-based crossword tasks.
Additionally, larger grids lead to decreasing performance, demonstrating the increased complexity of maintaining constraint adherence over a larger search space.


\textbf{LVLMs currently lag behind LLMs in performance, with minimal adherence to grid constraints.}
With image inputs, Claude 3.7 Sonnet achieves the highest performance among LVLMs,
but underperforms its own text inputs version.
Models such as GPT-4o and Gemini 2.0 Pro Exp exhibit similar trends, with WCRs below 0.35 on larger grids. All LVLMs demonstrate low ICRs, suggesting that they struggle to maintain reasoning consistency. Notably, performance declines when thinking mode is enabled on Claude 3.7 Sonnet with image input, which contradicts the improvements observed with text inputs; we explore this phenomenon in the next section. Among open-weight LVLMs, Pixtral-Large-Instruct achieves the best WCR of 0.297 on \texttt{7x7} grid, while still lags behind most proprietary LVLMs.

\subsection{Positive Correlation: LVLMs' Grid Parsing \& Puzzle-Solving Performance}\label{experiment: ocr}
\begin{wrapfigure}{r}{0.5\textwidth}
    \vspace{-1.0em}
    \centering
    \captionsetup{aboveskip=4pt, belowskip=4pt}
    \includesvg[width=0.5\textwidth]{graphs/ocr-performance-correlation.svg}
    \caption{Grid Parsing vs. Puzzle-Solving on \texttt{7×7} English puzzles, measured with WCR.}
    \label{fig: OCR_performance_correlation}
    \vspace{-2.0em}
\end{wrapfigure}

\textbf{Model performance on crossword puzzles exhibits a strong dependence on grid parsing capabilities, with systematic biases in word orientations.}
We evaluate grid parsing ability by prompting models to parse completed puzzle grids and associated clues (prompt details are in Appendix~\ref{appendix prompts}). Successful grid parsing requires: (1) identifying grid indexing numbers, (2) mapping numbers to clues to determine word orientation, and (3) extracting words with boundary recognition. 
The grid parsing ability reflects an LVLM's ability to interpret both spatial and textual information.
\begin{wraptable}{r}{0.5\textwidth} 
    \vspace{-1.3em}
    \centering
    \footnotesize
    \caption{WCR on Grid Parsing.}
    \begin{tabularx}{\linewidth}{@{}Xc|c}
        \toprule
        \textbf{Models} & \textbf{Across} & \textbf{Down} \\
        \midrule[\heavyrulewidth]
        \addlinespace[0.5em]
        \multicolumn{3}{c}{\cellcolor{customBlue}\textit{\textbf{Proprietary VLMs}}} \\
        \midrule
        
        \bf\mbox{Claude-3-7-Sonnet}                                             & \err{0.954}{0.009} & \err{0.760}{0.018} \\
        \bf\mbox{Claude-3-7-Sonnet \faLightbulb[regular]}                       & \err{0.949}{0.010} & \err{0.654}{0.022} \\
        \bf\mbox{GPT-4o-2024-11-20}                                             & \err{0.886}{0.014} & \err{0.448}{0.024} \\

        \midrule[\heavyrulewidth]
        \addlinespace[0.5em]
        \multicolumn{3}{c}{\cellcolor{customBlue}\textit{\textbf{Open-Weight VLMs}}} \\
        \midrule

        \bf\mbox{Pixtral-Large-Instruct}                    & \err{0.753}{0.022} & \err{0.361}{0.022} \\
        \bf\mbox{QVQ-72B-Preview}                           & \err{0.717}{0.022} & \err{0.139}{0.019} \\
        \bottomrule
    \end{tabularx}
    \vspace{-1.0em}
    \label{tab:small OCR results}
\end{wraptable}

We reveal two significant patterns:
(1) As shown in Table~\ref{tab:small OCR results}, LVLMs extract Across words more accurately than Down words, highlighting OCR limitations.
(2) Figure~\ref{fig: OCR_performance_correlation} demonstrates a strong positive correlation between grid parsing and overall puzzle-solving performance (both measured by WCR).
Notably, when enabling thinking mode, Claude 3.7 Sonnet exhibits lower grid parsing performance, which aligns with its lower puzzle-solving performance.
We hypothesize that the additional token generation in thinking mode diverts attention from image processing. 
These findings highlight the critical role of precise spatial-textual interpretation for future LVLM design. For full results, please refer to Appendix~\ref{grid_parsing_full}.

\subsection{Agentic Evaluation Setting: Interactive Mode for LVLMs}\label{experiment: interactive mode}
Motivated by recent research on visual-of-thoughts~\citep{wu2024mind, li2025imagine}, we introduce a new evaluation setting for LVLMs, referred to as \textbf{Interactive Mode}. This setting leverages the ability of \benchmarkname{} to automatically generate updated grid images. 

\textbf{Interactive Mode} requires step-by-step puzzle solving rather than completing the entire puzzle in a single pass. Specifically, the implementation of the controllable generation framework allows for updating grid images each time a model provides an  answer.
\begin{wrapfigure}{r}{0.5\textwidth}
    \centering
    \vspace{-1.0em}
    \includesvg[width=0.5\textwidth]{graphs/interactive-step-cdf.svg}
    \caption{CDF of ISS on \texttt{7x7} English Puzzles.}
    \label{fig: Interactive Metrics}
    \vspace{-1.5em}
\end{wrapfigure}
We maintain a four-round conversation history due to context window limitations. 
We introduce
\textbf{Interactive Success Step (ISS)} to quantify how many words a model correctly solves before making its first error. 
Figure~\ref{fig: Interactive Metrics} shows the cumulative distribution of ISS for each model---\textbf{most models fail at the initial solution step on most puzzles}.

Unlike the visual-of-thoughts approach, which focuses on generating intermediate reasoning plots, we employ external functions to update the grid. 
This establishes a foundation for evaluating LVLMs in agentic settings, where the word placement functions can serve as callable tools for the model and provide error feedback to guide the model in refining its responses---for example, by indicating whether a proposed word matches the expected length and whether its intersecting characters align with previously filled cells.

\subsection{Beyond English: Evaluations on Multilingual, Dictionary-based, and Adapted Data}\label{experiment_extension}
\begin{table}[htbp]
    \centering
    \footnotesize
    \caption{WCR and ICR for Chinese, English Simple, and CommonSenseQA puzzle sets.}
    \setlength{\tabcolsep}{2pt} 
    \renewcommand{\arraystretch}{1.3} 
    \resizebox{1.0\linewidth}{!}{
    \begin{tabularx}{\textwidth}{@{}Xcc|cc|cc}
        \toprule
        \multirow{2}{*}{\textbf{Models}} 
         & \multicolumn{2}{c}{\textbf{Chinese}} 
         & \multicolumn{2}{c}{\textbf{English Simple}} 
         & \multicolumn{2}{c}{\textbf{CommonSenseQA}} \\
        \cmidrule(lr){2-3} \cmidrule(lr){4-5} \cmidrule(lr){6-7}
         & \textbf{WCR} & \textbf{ICR} 
         & \textbf{WCR} & \textbf{ICR} 
         & \textbf{WCR} & \textbf{ICR} \\
        \midrule[\heavyrulewidth]
        \addlinespace[0.5em]
        \multicolumn{7}{c}{\cellcolor{customBlue}\textit{\textbf{Proprietary LVLMs}}} \\
        \midrule
        \bf\mbox{GPT-4o-2024-11-20} 
         & \err{0.366}{0.018} & \err{0.170}{0.017} 
         & \err{0.335}{0.015} & \err{0.227}{0.017} 
         & \err{0.392}{0.026} & \err{0.247}{0.026} \\
        \bf\mbox{Gemini 2.0 Flash}  
         & \err{0.208}{0.031} & \err{0.233}{0.018} 
         & \err{0.229}{0.014} & \err{0.216}{0.013} 
         & \err{0.327}{0.021} & \err{0.216}{0.018} \\
        
        \midrule[\heavyrulewidth]
        \addlinespace[0.5em]
        \multicolumn{7}{c}{\cellcolor{customBlue}\textit{\textbf{Open-Weight LVLMs}}} \\
        \midrule
        \bf\mbox{Pixtral-Large-Instruct-2411} 
         & \err{0.252}{0.015} & \err{0.101}{0.011} 
         & \err{0.216}{0.016} & \err{0.187}{0.015} 
         & \err{0.439}{0.023} & \err{0.270}{0.023} \\
        \bf\mbox{Qwen2.5-VL-72B-Instruct} 
         & \err{0.391}{0.025} & \err{0.282}{0.018} 
         & \err{0.239}{0.016} & \err{0.183}{0.015}  
         & \err{0.418}{0.022} & \err{0.252}{0.023}  \\
        
        \midrule[\heavyrulewidth]
        \addlinespace[0.5em]
        \multicolumn{7}{c}{\cellcolor{customGreen}\textit{\textbf{Proprietary LLMs}}} \\
        \midrule
        \bf\mbox{GPT-4o-2024-11-20} 
         & \err{0.593}{0.019} & \err{0.448}{0.021} 
         & \err{0.438}{0.020} & \err{0.306}{0.022} 
         & \err{0.524}{0.024} & \err{0.326}{0.029} \\
        \bf\mbox{o3-mini-high \faLightbulb[regular]} 
         & \err{0.774}{0.016} & \err{0.953}{0.014} 
         & \err{0.782}{0.021} & \err{0.946}{0.012} 
         & \err{0.812}{0.027} & \err{0.971}{0.011}  \\
        
        \midrule[\heavyrulewidth]
        \addlinespace[0.5em]
        \multicolumn{7}{c}{\cellcolor{customGreen}\textit{\textbf{Open-Weight LLMs}}} \\
        \midrule
        \bf\mbox{DeepSeek-R1 \faLightbulb[regular]} 
         & \err{0.907}{0.016} & \err{0.898}{0.018} 
         & \err{0.759}{0.017} & \err{0.787}{0.020}
         & \err{0.752}{0.031} & \err{0.829}{0.026}  \\
        \bf\mbox{QwQ-32B \faLightbulb[regular]} 
         & \err{0.701}{0.020} & \err{0.654}{0.022} 
         & \err{0.647}{0.021} & \err{0.734}{0.020} 
         & \err{0.699}{0.026} & \err{0.766}{0.027} \\
        \bottomrule
    \end{tabularx}}
    \label{tab:extension_results}
\end{table}

We extend zero-shot CoT prompting evaluation to Chinese, dictionary-based, and benchmark-adapted word–clue pairs in \benchmarkname{}, with results shown in Table~\ref{tab:extension_results}.

\textbf{Reasoning models outperform conventional ones across metrics, and the performance gap between open-weight and proprietary models is reduced.}
o3-mini maintains a high ICR across all three datasets, consistent with the main evaluation results. Furthermore, the performance gap between open-weight and proprietary reasoning models is reduced, possibly due to the simplicity of these tasks and differences in training data. Interestingly, when solving Chinese puzzles, we observe that QwQ-32B consistently reasons in Chinese.

\textbf{Effectiveness of grid construction: puzzles constructed with CommonsenseQA training data remain challenging.}
Despite CommonsenseQA having saturated performance on most LLMs, its training set continues to pose significant challenges for reasoning models. For example, o3-mini achieves the highest WCR of 0.812, yet it still falls short of perfection. In contrast, the best-performing non-reasoning LLMs such as GPT-4o and open-weight LVLMs Pixtral-Large-Instruct obtain WCR of 0.524 and 0.439, respectively. These results highlight the effectiveness of our grid construction strategy in repurposing existing benchmark data.

\vspace{-0.5em}
\section{Analysis}\label{Analysis}
\vspace{-0.5em}
In this section, we provide an analysis of model behavior on \benchmarkname{} through the impact of structural constraints and reasoning mechanisms, and discuss future applications.

\begin{wraptable}{r}{0.5\textwidth} 
    \vspace{-1.3em}
    \centering
    \footnotesize
    \caption{Grid Format Ablations.}
    \begin{tabularx}{\linewidth}{@{}Xc|c}
        \toprule
        \textbf{Models} & \textbf{Array} & \textbf{Markdown} \\
        \midrule 
        \bf\mbox{Claude-3-7-Sonnet}                                             & \err{0.482}{0.015} & \err{0.760}{0.018} \\
        \bf\mbox{GPT-4o-2024-11-20}                                             & \err{0.410}{0.014} & \err{0.398}{0.024} \\
        \bf\mbox{Gemini 2.0 Flash}                                              & \err{0.301}{0.014} & \err{0.309}{0.015} \\
        \bf\mbox{DeepSeek-V3}                                                   & \err{0.303}{0.014} & \err{0.294}{0.016} \\
        \bf\mbox{o3-mini}                                                       & \err{0.587}{0.023} & \err{0.592}{0.024} \\
        \bottomrule
    \end{tabularx}
    \vspace{-1.5em}
    \label{tab:grid format}
\end{wraptable}
\subsection{Grid Format Ablations: LLMs Robustness to Textual Grid Representations}\label{grid_format_ablation}

\textbf{LLMs exhibit robustness to variations in grid text representation.}
In the text-only evaluation setting, the empty crossword puzzle grid is represented as a 2D binary array, where \texttt{1} denotes a blocked cell and \texttt{0} denotes an unfilled cell. This array is then prepended to the prompt containing clues.
To evaluate the impact of different formatting choices, we test an alternative markdown-style representation in this section, with \texttt{·} indicates unfilled cells and \texttt{-} indicates blocked cells. An example of the two formats is provided in Appendix~\ref{appendix grid} and Figure~\ref{fig:grid_representations}.

Due to space constraints in the main text, we defer additional ablation studies on LVLM input formats and few-shot prompting to Appendix~\ref{appendix: LVLM format} and~\ref{appendix: few_shot}, respectively. We also demonstrate how controlling prefill ratio can be used to adjust task difficulty. We encourage readers to refer to these appendices for a comprehensive analysis and further discussion.

\subsection{Crossing Letters: Reasoning LLMs Improve with More Intersections}\label{experiment:intersection_analysis}
\begin{wrapfigure}{r}{0.5\textwidth}
    \centering
    \vspace{-1.3em}
    \includesvg[width=0.5\textwidth]{graphs/combined_average_accuracy_by_range_plot.svg}
    \caption{Crossing letter counts and average WCR on \texttt{7x7} English puzzles across LLMs.}
    \label{fig:intersection_accuracy}
    \vspace{-1.5em}
\end{wrapfigure}
\textbf{Average word accuracy on reasoning models increases with crossing letter count.}

\begin{wraptable}{r}{0.5\textwidth} 
    \vspace{-2.8em}
    \centering
    \footnotesize
     \caption{Average WCR results by crossing letter count and density on 7x7 English puzzles.}
     \vspace{-0.75em}
    \renewcommand{\arraystretch}{0.28}  
    \setlength{\tabcolsep}{3pt} 
    \begin{tabularx}{\linewidth}{@{}Xccc}
        \toprule
        \textbf{Density} & \textbf{Crossing Letter} & \textbf{Reason.} & \textbf{Non-Reason.} \\
        \midrule
        \multirow{3}{*}{Low}    & Low(1)   & 0.4341 & 0.2960 \\
                                & Med(2)   & 0.5275 & 0.3355 \\
                                & High(3+)  & 0.6165 & 0.2794 \\
        \midrule
        \multirow{3}{*}{Medium} & Low(1)   & 0.4271 & 0.2754 \\
                                & Med(2)   & 0.5477 & 0.3089 \\
                                & High(3+)  & 0.6029 & 0.4193 \\
        \midrule
        \multirow{3}{*}{High}   & Low(1)   & 0.5288 & 0.3097 \\
                                & Med(2)   & 0.5151 & 0.3217 \\
                                & High(3+) & 0.5928 & 0.3030 \\
        \bottomrule
    \end{tabularx}
    \captionsetup{aboveskip=4.0pt, belowskip=4.0pt}
    \vspace{-3.0em}
    \label{tab:crossing_letter}
\end{wraptable}

Figure~\ref{fig:intersection_accuracy} presents the average WCR for each range of crossing letter counts, divided into three groups, across five reasoning and nine non-reasoning LLMs.
We observe that reasoning models exhibit increasing accuracy with a greater number of letter intersections, whereas this trend is not observed for non-reasoning LLMs.
This observation aligns with our main experiment results, in which reasoning models tend to exhibit higher ICRs, suggesting that they benefit from effectively utilizing grid constraints for solution space reduction.
However, we do not observe the same pattern on 14x14 puzzles---likely because puzzles with larger grids are substantially more difficult even for reasoning models. We provide detailed results in Appendix~\ref{appendix_crossing_letter_constraints}.

To further isolate the effect of crossing-letter density within a constrained grid, we perform a quasi-ablation study on 7x7 puzzles by stratifying puzzles into low, medium, and high-density groups based on the average number of crossing letters per word, as shown in Table~\ref{tab:crossing_letter}.
Overall, within different densities, reasoning LLMs demonstrate improved WCR with a greater number of crossing letters, whereas non-reasoning LLMs show relatively stable performance.

\subsection{Self-Reflection: Limited Impact on Crossword Puzzle Solving}\label{experiment:self_reflection}
\begin{wrapfigure}{r}{0.5\textwidth}
    \centering
    \vspace{-1.5em}
    \includesvg[width=0.5\textwidth]{graphs/self_reflection_individual_radar_comparison.svg}
    \captionsetup{aboveskip=-0.1pt, belowskip=-0.1pt}
    \caption{Self-reflection improvements on \texttt{7x7} English puzzles across metrics. Top: reasoning LLMs. Bottom: non-reasoning LVLMs.}
    \label{fig: Self Reflection}
    \vspace{-2.0em}
\end{wrapfigure}
Backtracking and verifying previously filled answers are essential components of effective crossword puzzle solving.
To evaluate the effect of self-reflection, we include a manually crafted follow-up query that prompts models to revisit their previous zero-shot CoT responses (please see Appendix~\ref{appendix prompts} for details).
As shown in Figure~\ref{fig: Self Reflection}, we observe no measurable performance improvement for reasoning LLMs or non-reasoning LVLMs.
This find suggests that additional interactions with manual prompting alone are insufficient to enhance reasoning capabilities for puzzle solving.

\subsection{Test-Time Scaling: Diminishing Returns on Puzzle Performance}\label{experiment:TTS}
\begin{wrapfigure}{r}{0.5\textwidth}
    \centering
    \vspace{-1.3em}
    \captionsetup{aboveskip=-0.5pt, belowskip=-0.5pt}
    \includesvg[width=0.5\textwidth]{graphs/reasoning_effort_bar.svg}
    \caption{o3-mini performance on \texttt{7x7} English puzzles under three levels of reasoning efforts.}
    \label{fig:TTS}
    \vspace{-2.0em}
\end{wrapfigure}
To examine the effect of test-time scaling on CrossWordBench, we evaluate o3-mini on  \texttt{7x7} English puzzles by varying its reasoning effort, which controls the number of reasoning tokens generated during inference.
As shown in Figure~\ref{fig:TTS}, increasing the effort from low to medium yields a substantial performance improvement across all three metrics. However, further doubling the reasoning tokens provides no significant additional gains, indicating diminishing returns.

\subsection{Discussion: Verifiable Crossword for Future Multimodal RL Training}

Table~\ref{tab:main_results} shows that reasoning models such as Open-Reasoner-Zero-32B, which are primarily trained on mathematical problems, exhibit limited generalization to \benchmarkname{}.
This highlights the limitations of relying predominantly on math-based tasks with verifiable answers for reinforcement learning in more complex reasoning settings.
As multimodal reasoning advances, a significant challenge persists: the lack of multimodal environments with verifiable answers suitable for rule-based reinforcement learning.
We propose crossword puzzles as a compelling alternative, owing to their unique combinations of verifiability, multimodal structure, and goal-oriented task design. Crossword puzzles are also suitable for multi-turn training, where each word fill can represent a discrete, observable decision.

Our framework supports this setting by providing word placement functions that dynamically update the puzzle grid in both text and image formats, allowing seamless integration with multimodal agents and enabling tool-use training.
We leave the exploration of training multimodal agents on crossword puzzles with real-time function feedback for future work.

\vspace{-0.5em}
\section{Conclusion}\label{Conclusion}
\vspace{-0.5em}
This paper introduces \benchmarkname{}, a benchmark designed to evaluate the multimodal reasoning capabilities of both LLMs and LVLMs using crossword puzzles, which uniquely integrate text-based clues and visual constraints.
Our extensive evaluation of over 20 models shows that reasoning models substantially outperform non-reasoning counterparts and can benefit from increased crossing-letter constraints.
Additionally, we find a strong correlation between puzzle-solving performance and grid-parsing accuracy in LVLMs.
Even puzzles derived from saturated benchmarks remain challenging, emphasizing the necessity of structural complexity in rigorous reasoning evaluation.
This work paves the way for improving future multimodal RL training where interplay between modalities is essential.

\clearpage




\bibliography{colm2025_conference}

\begin{thebibliography}{62}
\providecommand{\natexlab}[1]{#1}
\providecommand{\url}[1]{\texttt{#1}}
\expandafter\ifx\csname urlstyle\endcsname\relax
  \providecommand{\doi}[1]{doi: #1}\else
  \providecommand{\doi}{doi: \begingroup \urlstyle{rm}\Url}\fi

\bibitem[Abdin et~al.(2024)Abdin, Aneja, Behl, Bubeck, Eldan, Gunasekar, Harrison, Hewett, Javaheripi, Kauffmann, et~al.]{abdin2024phi}
Marah Abdin, Jyoti Aneja, Harkirat Behl, S{\'e}bastien Bubeck, Ronen Eldan, Suriya Gunasekar, Michael Harrison, Russell~J Hewett, Mojan Javaheripi, Piero Kauffmann, et~al.
\newblock Phi-4 technical report.
\newblock \emph{arXiv preprint arXiv:2412.08905}, 2024.

\bibitem[Agrawal et~al.(2024)Agrawal, Antoniak, Hanna, Bout, Chaplot, Chudnovsky, Costa, De~Monicault, Garg, Gervet, et~al.]{agrawal2024pixtral}
Pravesh Agrawal, Szymon Antoniak, Emma~Bou Hanna, Baptiste Bout, Devendra Chaplot, Jessica Chudnovsky, Diogo Costa, Baudouin De~Monicault, Saurabh Garg, Theophile Gervet, et~al.
\newblock Pixtral 12b.
\newblock \emph{arXiv preprint arXiv:2410.07073}, 2024.

\bibitem[Ahn et~al.(2024)Ahn, Verma, Lou, Liu, Zhang, and Yin]{ahn2024large}
Janice Ahn, Rishu Verma, Renze Lou, Di~Liu, Rui Zhang, and Wenpeng Yin.
\newblock Large language models for mathematical reasoning: Progresses and challenges.
\newblock \emph{arXiv preprint arXiv:2402.00157}, 2024.

\bibitem[Anthropic(2024)]{anthropic_claude_2024}
Anthropic.
\newblock Claude 3.5 sonnet.
\newblock \url{https://www.anthropic.com/news/claude-3-5-sonnet}, 2024.

\bibitem[Austin et~al.(2021)Austin, Odena, Nye, Bosma, Michalewski, Dohan, Jiang, Cai, Terry, Le, and Sutton]{Austin2021ProgramSW}
Jacob Austin, Augustus Odena, Maxwell Nye, Maarten Bosma, Henryk Michalewski, David Dohan, Ellen Jiang, Carrie~J. Cai, Michael Terry, Quoc~V. Le, and Charles Sutton.
\newblock Program synthesis with large language models.
\newblock \emph{ArXiv}, abs/2108.07732, 2021.
\newblock URL \url{https://api.semanticscholar.org/CorpusID:237142385}.

\bibitem[Bai et~al.(2025)Bai, Chen, Liu, Wang, Ge, Song, Dang, Wang, Wang, Tang, et~al.]{bai2025qwen2}
Shuai Bai, Keqin Chen, Xuejing Liu, Jialin Wang, Wenbin Ge, Sibo Song, Kai Dang, Peng Wang, Shijie Wang, Jun Tang, et~al.
\newblock Qwen2. 5-vl technical report.
\newblock \emph{arXiv preprint arXiv:2502.13923}, 2025.

\bibitem[Besta et~al.(2024)Besta, Blach, Kubicek, Gerstenberger, Podstawski, Gianinazzi, Gajda, Lehmann, Niewiadomski, Nyczyk, et~al.]{besta2024graph}
Maciej Besta, Nils Blach, Ales Kubicek, Robert Gerstenberger, Michal Podstawski, Lukas Gianinazzi, Joanna Gajda, Tomasz Lehmann, Hubert Niewiadomski, Piotr Nyczyk, et~al.
\newblock Graph of thoughts: Solving elaborate problems with large language models.
\newblock In \emph{Proceedings of the AAAI Conference on Artificial Intelligence}, volume~38, pp.\  17682--17690, 2024.

\bibitem[Chen et~al.(2025)Chen, Wei, Ren, Li, and Zhang]{chen2025lr}
Jianghao Chen, Zhenlin Wei, Zhenjiang Ren, Ziyong Li, and Jiajun Zhang.
\newblock Lr$^2$bench: Evaluating long-chain reflective reasoning capabilities of large language models via constraint satisfaction problems, 2025.
\newblock URL \url{https://arxiv.org/abs/2502.17848}.

\bibitem[Chen et~al.(2022)Chen, Ma, Wang, and Cohen]{chen2022program}
Wenhu Chen, Xueguang Ma, Xinyi Wang, and William~W Cohen.
\newblock Program of thoughts prompting: Disentangling computation from reasoning for numerical reasoning tasks.
\newblock \emph{arXiv preprint arXiv:2211.12588}, 2022.

\bibitem[Chen et~al.(2024)Chen, Wu, Wang, Su, Chen, Xing, Zhong, Zhang, Zhu, Lu, et~al.]{chen2024internvl}
Zhe Chen, Jiannan Wu, Wenhai Wang, Weijie Su, Guo Chen, Sen Xing, Muyan Zhong, Qinglong Zhang, Xizhou Zhu, Lewei Lu, et~al.
\newblock Internvl: Scaling up vision foundation models and aligning for generic visual-linguistic tasks.
\newblock In \emph{Proceedings of the IEEE/CVF Conference on Computer Vision and Pattern Recognition}, pp.\  24185--24198, 2024.

\bibitem[Clark et~al.(2018)Clark, Cowhey, Etzioni, Khot, Sabharwal, Schoenick, and Tafjord]{allenai:arc}
Peter Clark, Isaac Cowhey, Oren Etzioni, Tushar Khot, Ashish Sabharwal, Carissa Schoenick, and Oyvind Tafjord.
\newblock Think you have solved question answering? try arc, the ai2 reasoning challenge.
\newblock \emph{ArXiv preprint}, abs/1803.05457, 2018.

\bibitem[Dai et~al.(2024)Dai, Lee, Wang, Yang, Liu, Barker, Rintamaki, Shoeybi, Catanzaro, and Ping]{nvlm2024}
Wenliang Dai, Nayeon Lee, Boxin Wang, Zhuolin Yang, Zihan Liu, Jon Barker, Tuomas Rintamaki, Mohammad Shoeybi, Bryan Catanzaro, and Wei Ping.
\newblock Nvlm: Open frontier-class multimodal llms.
\newblock \emph{arXiv preprint}, 2024.

\bibitem[DeepMind(2024)]{gemini_2}
Google DeepMind.
\newblock Gemini 2.0.
\newblock \url{https://deepmind.google/technologies/gemini}, 2024.

\bibitem[Devlin et~al.(2019)Devlin, Chang, Lee, and Toutanova]{devlin2019bert}
Jacob Devlin, Ming-Wei Chang, Kenton Lee, and Kristina Toutanova.
\newblock Bert: Pre-training of deep bidirectional transformers for language understanding.
\newblock In \emph{Proceedings of the 2019 conference of the North American chapter of the association for computational linguistics: human language technologies, volume 1 (long and short papers)}, pp.\  4171--4186, 2019.

\bibitem[Dubey et~al.(2024)Dubey, Jauhri, Pandey, Kadian, Al-Dahle, Letman, Mathur, Schelten, Yang, Fan, et~al.]{dubey2024llama}
Abhimanyu Dubey, Abhinav Jauhri, Abhinav Pandey, Abhishek Kadian, Ahmad Al-Dahle, Aiesha Letman, Akhil Mathur, Alan Schelten, Amy Yang, Angela Fan, et~al.
\newblock The llama 3 herd of models.
\newblock \emph{arXiv preprint arXiv:2407.21783}, 2024.

\bibitem[Efrat et~al.(2021)Efrat, Shaham, Kilman, and Levy]{efrat2021cryptonite}
Avia Efrat, Uri Shaham, Dan Kilman, and Omer Levy.
\newblock Cryptonite: A cryptic crossword benchmark for extreme ambiguity in language.
\newblock \emph{arXiv preprint arXiv:2103.01242}, 2021.

\bibitem[Feng et~al.(2023)Feng, Wan, Wen, McAleer, Wen, Zhang, and Wang]{feng2023alphazero}
Xidong Feng, Ziyu Wan, Muning Wen, Stephen~Marcus McAleer, Ying Wen, Weinan Zhang, and Jun Wang.
\newblock Alphazero-like tree-search can guide large language model decoding and training.
\newblock \emph{arXiv preprint arXiv:2309.17179}, 2023.

\bibitem[Ginsberg(2011)]{ginsberg2011dr}
Matthew~L Ginsberg.
\newblock Dr. fill: Crosswords and an implemented solver for singly weighted csps.
\newblock \emph{Journal of Artificial Intelligence Research}, 42:\penalty0 851--886, 2011.

\bibitem[Guo et~al.(2025)Guo, Yang, Zhang, Song, Zhang, Xu, Zhu, Ma, Wang, Bi, et~al.]{guo2025deepseek}
Daya Guo, Dejian Yang, Haowei Zhang, Junxiao Song, Ruoyu Zhang, Runxin Xu, Qihao Zhu, Shirong Ma, Peiyi Wang, Xiao Bi, et~al.
\newblock Deepseek-r1: Incentivizing reasoning capability in llms via reinforcement learning.
\newblock \emph{arXiv preprint arXiv:2501.12948}, 2025.

\bibitem[Huang et~al.(2025{\natexlab{a}})Huang, Huang, Leng, Liu, and Huang]{huang2025efficient}
Chengsong Huang, Langlin Huang, Jixuan Leng, Jiacheng Liu, and Jiaxin Huang.
\newblock Efficient test-time scaling via self-calibration.
\newblock \emph{arXiv preprint arXiv:2503.00031}, 2025{\natexlab{a}}.

\bibitem[Huang et~al.(2022)Huang, Gu, Hou, Wu, Wang, Yu, and Han]{huang2022large}
Jiaxin Huang, Shixiang~Shane Gu, Le~Hou, Yuexin Wu, Xuezhi Wang, Hongkun Yu, and Jiawei Han.
\newblock Large language models can self-improve.
\newblock \emph{arXiv preprint arXiv:2210.11610}, 2022.

\bibitem[Huang \& Chang(2022)Huang and Chang]{huang2022towards}
Jie Huang and Kevin Chen-Chuan Chang.
\newblock Towards reasoning in large language models: A survey.
\newblock \emph{arXiv preprint arXiv:2212.10403}, 2022.

\bibitem[Huang et~al.(2025{\natexlab{b}})Huang, Jia, Zhai, Cao, Ye, Zhao, Hu, and Lin]{huang2025vision}
Wenxuan Huang, Bohan Jia, Zijie Zhai, Shaosheng Cao, Zheyu Ye, Fei Zhao, Yao Hu, and Shaohui Lin.
\newblock Vision-r1: Incentivizing reasoning capability in multimodal large language models.
\newblock \emph{arXiv preprint arXiv:2503.06749}, 2025{\natexlab{b}}.

\bibitem[Hurst et~al.(2024)Hurst, Lerer, Goucher, Perelman, Ramesh, Clark, Ostrow, Welihinda, Hayes, Radford, et~al.]{hurst2024gpt}
Aaron Hurst, Adam Lerer, Adam~P Goucher, Adam Perelman, Aditya Ramesh, Aidan Clark, AJ~Ostrow, Akila Welihinda, Alan Hayes, Alec Radford, et~al.
\newblock Gpt-4o system card.
\newblock \emph{arXiv preprint arXiv:2410.21276}, 2024.

\bibitem[Jaech et~al.(2024)Jaech, Kalai, Lerer, Richardson, El-Kishky, Low, Helyar, Madry, Beutel, Carney, et~al.]{jaech2024openai}
Aaron Jaech, Adam Kalai, Adam Lerer, Adam Richardson, Ahmed El-Kishky, Aiden Low, Alec Helyar, Aleksander Madry, Alex Beutel, Alex Carney, et~al.
\newblock Openai o1 system card.
\newblock \emph{arXiv preprint arXiv:2412.16720}, 2024.

\bibitem[Kojima et~al.(2022)Kojima, Gu, Reid, Matsuo, and Iwasawa]{kojima2022large}
Takeshi Kojima, Shixiang~Shane Gu, Machel Reid, Yutaka Matsuo, and Yusuke Iwasawa.
\newblock Large language models are zero-shot reasoners.
\newblock \emph{Advances in neural information processing systems}, 35:\penalty0 22199--22213, 2022.

\bibitem[Kulshreshtha et~al.(2022)Kulshreshtha, Kovaleva, Shivagunde, and Rumshisky]{kulshreshtha2022down}
Saurabh Kulshreshtha, Olga Kovaleva, Namrata Shivagunde, and Anna Rumshisky.
\newblock Down and across: Introducing crossword-solving as a new nlp benchmark.
\newblock \emph{arXiv preprint arXiv:2205.10442}, 2022.

\bibitem[Li et~al.(2024{\natexlab{a}})Li, Zhang, Guo, Zhang, Li, Zhang, Zhang, Zhang, Li, Liu, et~al.]{li2024llava}
Bo~Li, Yuanhan Zhang, Dong Guo, Renrui Zhang, Feng Li, Hao Zhang, Kaichen Zhang, Peiyuan Zhang, Yanwei Li, Ziwei Liu, et~al.
\newblock Llava-onevision: Easy visual task transfer.
\newblock \emph{arXiv preprint arXiv:2408.03326}, 2024{\natexlab{a}}.

\bibitem[Li et~al.(2025)Li, Wu, Zhang, Xia, Mao, Dong, Vuli{\'c}, and Wei]{li2025imagine}
Chengzu Li, Wenshan Wu, Huanyu Zhang, Yan Xia, Shaoguang Mao, Li~Dong, Ivan Vuli{\'c}, and Furu Wei.
\newblock Imagine while reasoning in space: Multimodal visualization-of-thought.
\newblock \emph{arXiv preprint arXiv:2501.07542}, 2025.

\bibitem[Li et~al.(2024{\natexlab{b}})Li, Liu, Wu, Wang, Shen, Qu, Niu, Wang, Chen, and Li]{aria}
Dongxu Li, Yudong Liu, Haoning Wu, Yue Wang, Zhiqi Shen, Bowen Qu, Xinyao Niu, Guoyin Wang, Bei Chen, and Junnan Li.
\newblock Aria: An open multimodal native mixture-of-experts model.
\newblock \emph{arXiv preprint arXiv:2410.05993}, 2024{\natexlab{b}}.

\bibitem[Littman et~al.(2002)Littman, Keim, and Shazeer]{littman2002probabilistic}
Michael~L Littman, Greg~A Keim, and Noam Shazeer.
\newblock A probabilistic approach to solving crossword puzzles.
\newblock \emph{Artificial Intelligence}, 134\penalty0 (1-2):\penalty0 23--55, 2002.

\bibitem[Liu et~al.(2024)Liu, Feng, Xue, Wang, Wu, Lu, Zhao, Deng, Zhang, Ruan, et~al.]{liu2024deepseek}
Aixin Liu, Bei Feng, Bing Xue, Bingxuan Wang, Bochao Wu, Chengda Lu, Chenggang Zhao, Chengqi Deng, Chenyu Zhang, Chong Ruan, et~al.
\newblock Deepseek-v3 technical report.
\newblock \emph{arXiv preprint arXiv:2412.19437}, 2024.

\bibitem[Lu et~al.(2023)Lu, Bansal, Xia, Liu, Li, Hajishirzi, Cheng, Chang, Galley, and Gao]{lu2023mathvista}
Pan Lu, Hritik Bansal, Tony Xia, Jiacheng Liu, Chunyuan Li, Hannaneh Hajishirzi, Hao Cheng, Kai-Wei Chang, Michel Galley, and Jianfeng Gao.
\newblock Mathvista: Evaluating mathematical reasoning of foundation models in visual contexts.
\newblock \emph{arXiv preprint arXiv:2310.02255}, 2023.

\bibitem[Mirzadeh et~al.(2024)Mirzadeh, Alizadeh, Shahrokhi, Tuzel, Bengio, and Farajtabar]{mirzadeh2024gsm}
Iman Mirzadeh, Keivan Alizadeh, Hooman Shahrokhi, Oncel Tuzel, Samy Bengio, and Mehrdad Farajtabar.
\newblock Gsm-symbolic: Understanding the limitations of mathematical reasoning in large language models.
\newblock \emph{arXiv preprint arXiv:2410.05229}, 2024.

\bibitem[Muennighoff et~al.(2025)Muennighoff, Yang, Shi, Li, Fei-Fei, Hajishirzi, Zettlemoyer, Liang, Cand{\`e}s, and Hashimoto]{muennighoff2025s1}
Niklas Muennighoff, Zitong Yang, Weijia Shi, Xiang~Lisa Li, Li~Fei-Fei, Hannaneh Hajishirzi, Luke Zettlemoyer, Percy Liang, Emmanuel Cand{\`e}s, and Tatsunori Hashimoto.
\newblock s1: Simple test-time scaling.
\newblock \emph{arXiv preprint arXiv:2501.19393}, 2025.

\bibitem[OpenAI(2025)]{o3_mini}
OpenAI.
\newblock Openai o3-mini.
\newblock \url{https://openai.com/index/openai-o3-mini}, 2025.

\bibitem[Plaat et~al.(2024)Plaat, Wong, Verberne, Broekens, van Stein, and Back]{plaat2024reasoning}
Aske Plaat, Annie Wong, Suzan Verberne, Joost Broekens, Niki van Stein, and Thomas Back.
\newblock Reasoning with large language models, a survey.
\newblock \emph{arXiv preprint arXiv:2407.11511}, 2024.

\bibitem[Rozner et~al.(2021)Rozner, Potts, and Mahowald]{rozner2021decrypting}
Josh Rozner, Christopher Potts, and Kyle Mahowald.
\newblock Decrypting cryptic crosswords: Semantically complex wordplay puzzles as a target for nlp.
\newblock \emph{Advances in Neural Information Processing Systems}, 34:\penalty0 11409--11421, 2021.

\bibitem[Sadallah et~al.(2024)Sadallah, Kotova, and Kochmar]{sadallah2024llms}
Abdelrahman Sadallah, Daria Kotova, and Ekaterina Kochmar.
\newblock Are llms good cryptic crossword solvers?
\newblock \emph{arXiv preprint arXiv:2403.12094}, 2024.

\bibitem[Saha et~al.(2024)Saha, Chakraborty, Saha, and Garain]{saha2024language}
Soumadeep Saha, Sutanoya Chakraborty, Saptarshi Saha, and Utpal Garain.
\newblock Language models are crossword solvers.
\newblock \emph{arXiv preprint arXiv:2406.09043}, 2024.

\bibitem[Setlur et~al.(2024)Setlur, Nagpal, Fisch, Geng, Eisenstein, Agarwal, Agarwal, Berant, and Kumar]{setlur2024rewarding}
Amrith Setlur, Chirag Nagpal, Adam Fisch, Xinyang Geng, Jacob Eisenstein, Rishabh Agarwal, Alekh Agarwal, Jonathan Berant, and Aviral Kumar.
\newblock Rewarding progress: Scaling automated process verifiers for llm reasoning.
\newblock \emph{arXiv preprint arXiv:2410.08146}, 2024.

\bibitem[Talmor et~al.(2018)Talmor, Herzig, Lourie, and Berant]{talmor2018commonsenseqa}
Alon Talmor, Jonathan Herzig, Nicholas Lourie, and Jonathan Berant.
\newblock Commonsenseqa: A question answering challenge targeting commonsense knowledge.
\newblock \emph{arXiv preprint arXiv:1811.00937}, 2018.

\bibitem[Team et~al.(2025)Team, Kamath, Ferret, Pathak, Vieillard, Merhej, Perrin, Matejovicova, Ram{\'e}, Rivi{\`e}re, et~al.]{gemma_2025}
Gemma Team, Aishwarya Kamath, Johan Ferret, Shreya Pathak, Nino Vieillard, Ramona Merhej, Sarah Perrin, Tatiana Matejovicova, Alexandre Ram{\'e}, Morgane Rivi{\`e}re, et~al.
\newblock Gemma 3 technical report.
\newblock \emph{arXiv preprint arXiv:2503.19786}, 2025.

\bibitem[Team(2024)]{qvq-72b-preview}
Qwen Team.
\newblock Qvq: To see the world with wisdom, December 2024.
\newblock URL \url{https://qwenlm.github.io/blog/qvq-72b-preview/}.

\bibitem[Team(2025)]{qwq32b}
Qwen Team.
\newblock Qwq-32b: Embracing the power of reinforcement learning, March 2025.
\newblock URL \url{https://qwenlm.github.io/blog/qwq-32b/}.

\bibitem[Uesato et~al.(2022)Uesato, Kushman, Kumar, Song, Siegel, Wang, Creswell, Irving, and Higgins]{uesato2022solving}
Jonathan Uesato, Nate Kushman, Ramana Kumar, Francis Song, Noah Siegel, Lisa Wang, Antonia Creswell, Geoffrey Irving, and Irina Higgins.
\newblock Solving math word problems with process-and outcome-based feedback.
\newblock \emph{arXiv preprint arXiv:2211.14275}, 2022.

\bibitem[Veeraboina(2023)]{aime_1983_2024}
Hemish Veeraboina.
\newblock Aime problem set 1983-2024, 2023.
\newblock URL \url{https://www.kaggle.com/datasets/hemishveeraboina/aime-problem-set-1983-2024}.

\bibitem[Wallace et~al.(2022)Wallace, Tomlin, Xu, Yang, Pathak, Ginsberg, and Klein]{wallace2022automated}
Eric Wallace, Nicholas Tomlin, Albert Xu, Kevin Yang, Eshaan Pathak, Matthew Ginsberg, and Dan Klein.
\newblock Automated crossword solving.
\newblock \emph{arXiv preprint arXiv:2205.09665}, 2022.

\bibitem[Wang et~al.(2024)Wang, Pan, Shi, Lu, Ren, Zhou, Zhan, and Li]{wang2024measuring}
Ke~Wang, Junting Pan, Weikang Shi, Zimu Lu, Houxing Ren, Aojun Zhou, Mingjie Zhan, and Hongsheng Li.
\newblock Measuring multimodal mathematical reasoning with math-vision dataset.
\newblock \emph{Advances in Neural Information Processing Systems}, 37:\penalty0 95095--95169, 2024.

\bibitem[Wang et~al.(2023)Wang, Li, Shao, Xu, Dai, Li, Chen, Wu, and Sui]{wang2023math}
Peiyi Wang, Lei Li, Zhihong Shao, RX~Xu, Damai Dai, Yifei Li, Deli Chen, Y~Wu, and Zhifang Sui.
\newblock Math-shepherd: A label-free step-by-step verifier for llms in mathematical reasoning.
\newblock \emph{arXiv preprint arXiv:2312.08935}, 2023.

\bibitem[Wei et~al.(2022)Wei, Wang, Schuurmans, Bosma, Xia, Chi, Le, Zhou, et~al.]{wei2022chain}
Jason Wei, Xuezhi Wang, Dale Schuurmans, Maarten Bosma, Fei Xia, Ed~Chi, Quoc~V Le, Denny Zhou, et~al.
\newblock Chain-of-thought prompting elicits reasoning in large language models.
\newblock \emph{Advances in neural information processing systems}, 35:\penalty0 24824--24837, 2022.

\bibitem[Wu et~al.(2024)Wu, Mao, Zhang, Xia, Dong, Cui, and Wei]{wu2024mind}
Wenshan Wu, Shaoguang Mao, Yadong Zhang, Yan Xia, Li~Dong, Lei Cui, and Furu Wei.
\newblock Mind's eye of llms: Visualization-of-thought elicits spatial reasoning in large language models.
\newblock In \emph{The Thirty-eighth Annual Conference on Neural Information Processing Systems}, 2024.

\bibitem[Xu et~al.(2024)Xu, Jin, Hao, Song, Sun, and Yuan]{xu2024llava}
Guowei Xu, Peng Jin, Li~Hao, Yibing Song, Lichao Sun, and Li~Yuan.
\newblock Llava-o1: Let vision language models reason step-by-step.
\newblock \emph{arXiv preprint arXiv:2411.10440}, 2024.

\bibitem[Yang et~al.(2025)Yang, He, Pan, Jiang, Deng, Yang, Lu, Yin, Rao, Zhu, et~al.]{yang2025r1}
Yi~Yang, Xiaoxuan He, Hongkun Pan, Xiyan Jiang, Yan Deng, Xingtao Yang, Haoyu Lu, Dacheng Yin, Fengyun Rao, Minfeng Zhu, et~al.
\newblock R1-onevision: Advancing generalized multimodal reasoning through cross-modal formalization.
\newblock \emph{arXiv preprint arXiv:2503.10615}, 2025.

\bibitem[Yao et~al.(2023)Yao, Yu, Zhao, Shafran, Griffiths, Cao, and Narasimhan]{yao2023tree}
Shunyu Yao, Dian Yu, Jeffrey Zhao, Izhak Shafran, Tom Griffiths, Yuan Cao, and Karthik Narasimhan.
\newblock Tree of thoughts: Deliberate problem solving with large language models.
\newblock \emph{Advances in neural information processing systems}, 36:\penalty0 11809--11822, 2023.

\bibitem[Yao et~al.(2024)Yao, Yu, Zhang, Wang, Cui, Zhu, Cai, Li, Zhao, He, et~al.]{yao2024minicpm}
Yuan Yao, Tianyu Yu, Ao~Zhang, Chongyi Wang, Junbo Cui, Hongji Zhu, Tianchi Cai, Haoyu Li, Weilin Zhao, Zhihui He, et~al.
\newblock Minicpm-v: A gpt-4v level mllm on your phone.
\newblock \emph{arXiv preprint arXiv:2408.01800}, 2024.

\bibitem[Ye et~al.(2025)Ye, Huang, Xiao, Chern, Xia, and Liu]{ye2025limo}
Yixin Ye, Zhen Huang, Yang Xiao, Ethan Chern, Shijie Xia, and Pengfei Liu.
\newblock Limo: Less is more for reasoning.
\newblock \emph{arXiv preprint arXiv:2502.03387}, 2025.

\bibitem[Yue et~al.(2023)Yue, Ni, Zhang, Zheng, Liu, Zhang, Stevens, Jiang, Ren, Sun, Wei, Yu, Yuan, Sun, Yin, Zheng, Yang, Liu, Huang, Sun, Su, and Chen]{Yue2023MMMUAM}
Xiang Yue, Yuansheng Ni, Kai Zhang, Tianyu Zheng, Ruoqi Liu, Ge~Zhang, Samuel Stevens, Dongfu Jiang, Weiming Ren, Yuxuan Sun, Cong Wei, Botao Yu, Ruibin Yuan, Renliang Sun, Ming Yin, Boyuan Zheng, Zhenzhu Yang, Yibo Liu, Wenhao Huang, Huan Sun, Yu~Su, and Wenhu Chen.
\newblock Mmmu: A massive multi-discipline multimodal understanding and reasoning benchmark for expert agi.
\newblock \emph{2024 IEEE/CVF Conference on Computer Vision and Pattern Recognition (CVPR)}, pp.\  9556--9567, 2023.
\newblock URL \url{https://api.semanticscholar.org/CorpusID:265466525}.

\bibitem[Zeinalipour et~al.(2025)Zeinalipour, Saad, Maggini, and Gori]{zeinalipour2025arabic}
Kamyar Zeinalipour, Mohamed~Zaky Saad, Marco Maggini, and Marco Gori.
\newblock From arabic text to puzzles: Llm-driven development of arabic educational crosswords.
\newblock \emph{arXiv preprint arXiv:2501.11035}, 2025.

\bibitem[Zhang et~al.(2024{\natexlab{a}})Zhang, Zhoubian, Hu, Yue, Dong, and Tang]{zhang2024rest}
Dan Zhang, Sining Zhoubian, Ziniu Hu, Yisong Yue, Yuxiao Dong, and Jie Tang.
\newblock Rest-mcts*: Llm self-training via process reward guided tree search.
\newblock \emph{Advances in Neural Information Processing Systems}, 37:\penalty0 64735--64772, 2024{\natexlab{a}}.

\bibitem[Zhang et~al.(2024{\natexlab{b}})Zhang, Jiang, Zhang, Lin, Guo, Qiu, Zhou, Lu, Chang, Qiao, et~al.]{zhang2024mathverse}
Renrui Zhang, Dongzhi Jiang, Yichi Zhang, Haokun Lin, Ziyu Guo, Pengshuo Qiu, Aojun Zhou, Pan Lu, Kai-Wei Chang, Yu~Qiao, et~al.
\newblock Mathverse: Does your multi-modal llm truly see the diagrams in visual math problems?
\newblock In \emph{European Conference on Computer Vision}, pp.\  169--186. Springer, 2024{\natexlab{b}}.

\bibitem[Zhao et~al.(2025)Zhao, Wang, Peng, Zhao, Tian, Chen, Ji, and Li]{zhao202514millionopensourcedistilled}
Han Zhao, Haotian Wang, Yiping Peng, Sitong Zhao, Xiaoyu Tian, Shuaiting Chen, Yunjie Ji, and Xiangang Li.
\newblock 1.4 million open-source distilled reasoning dataset to empower large language model training, 2025.
\newblock URL \url{https://arxiv.org/abs/2503.19633}.

\end{thebibliography}
\bibliographystyle{colm2025_conference}

\clearpage
\appendix
\section{Acknowledgment} 
We thank all reviewers for their feedback, which contributed to improving this work.

\section{Limitations}
We propose CrossWordBench to evaluate the reasoning capabilities of both LLMs and LVLMs. 
Unlike prior work, CrossWordBench does not incorporate any human-authored puzzles from online sources.
Although our generation process is carefully designed to preserve key structural properties—such as clue distribution, word length, and blocked cell ratios---as summarized in Table~\ref{tab:crossword_stats_all}, and the generated puzzles exhibit difficulty levels comparable to those of human---created puzzles when evaluated using standard word-level metrics, manual verification of puzzle quality could be helpful. Although our framework is also capable of generating crossword puzzles for potential use in training, this paper focuses solely on evaluation and leaves the exploration of reinforcement learning for future work.

\section{More Results and Analysis}
\subsection{Behavior Analysis}
To better analyze the errors made by models in puzzle-solving, we define two metrics: global length error and local length error.
The global length error metric compares the number of words produced by the model with those in the reference answer list, assessing whether the model supplies an answer for every clue in the puzzle.
In contrast, the local length error metric compares the length of each individual word to its corresponding reference, thereby quantifying the model's adherence to the grid constraints.
Table~\ref{tab:length_results} shows that even the best-performing reasoning models, such as Claude 3.7 Sonnet with thinking mode and DeepSeek R1, exhibit global length errors on two puzzles.
Almost all models---with the exception of o3-mini, which demonstrates the highest ICR---commit a significant number of local errors.
Moreover, we observe that all global length errors arise from models providing fewer answers than required, i.e., failing to address some clues.
In contrast, most local length errors are due to models generating answers that exceed the expected word length.
\begin{table}[htbp]
    \centering
    \setlength{\tabcolsep}{2pt} 
    \renewcommand{\arraystretch}{1.3} 
    \footnotesize
    \caption{Global and Local Length Errors across models on \texttt{7x7} English puzzles. ``Long" and ``Short" indicate words that are longer or shorter than the corresponding reference answer.}
    \resizebox{\linewidth}{!}{
    \begin{tabularx}{\textwidth}{@{}Xccc|ccc}
        \toprule
        \multirow{2}{*}{\textbf{Models}} & \multicolumn{3}{c}{\textbf{Global Length Error}} & \multicolumn{3}{c}{\textbf{Local Length Error}} \\
        \cmidrule(lr){2-4} \cmidrule(lr){5-7}
                        & \textbf{Tot.} & \textbf{Long} & \textbf{Short} 
                        & \textbf{Tot.} & \textbf{Long} & \textbf{Short} \\
        \midrule[\heavyrulewidth]
        \addlinespace[0.5em]
        \multicolumn{7}{c}{\cellcolor{customBlue}\textit{\textbf{Proprietary LVLMs}}} \\
        \midrule
        
        \bf\mbox{Claude-3-7-Sonnet}                                                 & 0 & 0 & 0 & 363 & 205 & 158 \\
        
        \bf\mbox{Claude-3-7-Sonnet \faLightbulb[regular]}                           & 0 & 0 & 0 & 454 & 237 & 217 \\
        
        \bf\mbox{GPT-4o-2024-11-20}                                                 & 10 & 0 & 10 & 581 & 326 & 255 \\
        
        \bf\mbox{Gemini 2.0 Pro Exp}                                                & 1 & 0 & 1 & 565 & 467 & 98 \\
        
        \bf\mbox{Gemini 2.0 Flash}                                                  & 0 & 0 & 0 & 665 & 568 & 97 \\

        \midrule[\heavyrulewidth]
        \addlinespace[0.5em]
        \multicolumn{7}{c}{\cellcolor{customBlue}\textit{\textbf{Open-Weight LVLMs}}} \\
        \midrule
        \bf\mbox{Pixtral-Large-Instruct-2411}                                   & 3 & 0 & 3 & 623 & 481 & 142 \\
        
        \bf\mbox{InternVL2\_5-78B-MPO}                                          & 0 & 0 & 0 & 834 & 682 & 152 \\

        \bf\mbox{NVLM-D-72B}                                                    & 8 & 0 & 8 & 791 & 620 & 171 \\
        
        \bf\mbox{Qwen2.5-VL-72B-Instruct}                                       & 2 & 0 & 2 & 744 & 600 & 144 \\
        
        \bf\mbox{QVQ-72B-Preview}                                               & 20 & 0 & 20 & 765 & 525 & 240 \\
        
        \bf\mbox{llava-onevision-72b-ov-chat}                                   & 3 & 0 & 3 & 829 & 715 & 114 \\

        \bf\mbox{gemma-3-27b-it}                                                & 18 & 0 & 18 & 781 & 645 & 136 \\
        
        \bf\mbox{Aria}                                                          & 16 & 0 & 16 & 894 & 720 & 174 \\
        
        \bf\mbox{MiniCPM-V-2\_6}                                                & 16 & 0 & 16 & 918 & 688 & 230 \\
        
        \bf\mbox{Qwen2.5-VL-3B-Instruct}                                        & 31 & 0 & 31 & 1034 & 613 & 421 \\
        
        \midrule[\heavyrulewidth]
        \addlinespace[0.5em]
        \multicolumn{7}{c}{\cellcolor{customGreen}\textit{\textbf{Proprietary LLMs}}} \\
        \midrule
        \bf\mbox{o3-mini} \faLightbulb[regular]                                     & 0 & 0 & 0 & 6 & 4 & 2 \\

        \bf\mbox{Claude-3-7-Sonnet \faLightbulb[regular]}                           & 2 & 0 & 2 & 124 & 44 & 80 \\

        \bf\mbox{Claude-3-7-Sonnet}                                                 & 1 & 0 & 1 & 274 & 74 & 200 \\
        
        \bf\mbox{GPT-4o-2024-11-20}                                                 & 17 & 0 & 17 & 399 & 183 & 216 \\
        
        \bf\mbox{Gemini 2.0 Pro Exp}                                                & 0 & 0 & 0 & 378 & 254 & 124 \\
        
        \bf\mbox{Gemini 2.0 Flash}                                                  & 0 & 0 & 0 & 633 & 545 & 88 \\
        
        \midrule[\heavyrulewidth]
        \addlinespace[0.5em]
        \multicolumn{7}{c}{\cellcolor{customGreen}\textit{\textbf{Open-Weight LLMs}}} \\

        \bf\mbox{Llama-3.1-405B-Instruct}$^{\dagger}$                           & 8 & 0 & 8 & 835 & 741 & 94 \\
        
        \bf\mbox{DeepSeek-R1 \faLightbulb[regular]}                             & 2 & 0 & 2 & 25 & 14 & 11 \\
        
        \bf\mbox{DeepSeek-V3}$^{\dagger}$                                       & 11 & 0 & 11 & 513 & 206 & 307 \\
        
        \bf\mbox{R1-Distill-Llama-70B \faLightbulb[regular]}                    & 9 & 0 & 9 & 203 & 110 & 93 \\
        
        \bf\mbox{Llama-3.3-70B-Instruct}                                        & 22 & 0 & 22 & 598 & 326 & 272 \\
        
        \bf\mbox{QwQ-32B \faLightbulb[regular] }                                & 9 & 0 & 9 & 65 & 34 & 31 \\
        
        \bf\mbox{Open-Reasoner-Zero-32B \faLightbulb[regular]}                  & 10 & 0 & 10 & 697 & 473 & 224 \\
            
        \bf\mbox{Phi-4}                                                         & 2 & 0 & 2 & 709 & 447 & 262 \\
        
        \bottomrule
    \end{tabularx}}
    \label{tab:length_results}
\end{table}

\subsection{Complete Results on Grid Parsing Performance}\label{grid_parsing_full}
In Section~\ref{experiment: ocr}, we demonstrate a positive correlation between LVLM's grid-parsing accuracy and their puzzle-solving performance. 
We also provide several examples illustrating the limitations of LVLMs in extracting horizontal words.
Here, we present a complete table of all models shown in Table~\ref{fig: OCR_performance_correlation}, as summarized in Table~\ref{tab:OCR results}, where a similar trend is observed.
\begin{table}[htbp] 
    \centering
    \footnotesize
        \caption{WCR of Grid Parsing for all models.}
    \begin{tabularx}{\textwidth}{@{}Xcc|c|c}
        \toprule
        \textbf{Models} & \textbf{Size} & \textbf{Across} & \textbf{Down} & \textbf{All} \\
        \midrule[\heavyrulewidth]
        \addlinespace[0.5em]
        \multicolumn{5}{c}{\cellcolor{customBlue}\textit{\textbf{Proprietary LVLMs}}} \\
        \midrule

        \bf\mbox{Claude-3-7-Sonnet}                                             & -             & \err{0.954}{0.009} & \err{0.760}{0.018} & \err{0.855}{0.010} \\
        \bf\mbox{Claude-3-7-Sonnet} \faLightbulb[regular]                       & -             & \err{0.949}{0.010} & \err{0.654}{0.022} & \err{0.800}{0.012} \\

        \bf\mbox{GPT-4o-2024-11-20}                                             & -             & \err{0.886}{0.014} & \err{0.448}{0.024} & \err{0.668}{0.015} \\
        \bf\mbox{Gemini 2.0 Pro Exp}                                            & -             & \err{0.962}{0.008} & \err{0.693}{0.018} & \err{0.826}{0.011} \\
        \bf\mbox{Gemini 2.0 Flash}                                              & -             & \err{0.954}{0.009} & \err{0.381}{0.024} & \err{0.667}{0.013} \\
        
        \midrule[\heavyrulewidth]
        \addlinespace[0.5em]
        \multicolumn{5}{c}{\cellcolor{customBlue}\textit{\textbf{Open-Weight LVLMs}}} \\
        \midrule

        \bf\mbox{Pixtral-Large-Instruct-2411}               & 124B      & \err{0.753}{0.022} & \err{0.361}{0.022} & \err{0.556}{0.016} \\
        \bf\mbox{NVLM-D-72B}                                & 78.4B     & \err{0.429}{0.024} & \err{0.099}{0.015} & \err{0.261}{0.013} \\
        \bf\mbox{InternVL2\_5-78B-MPO}                      & 78.4B     & \err{0.744}{0.019} & \err{0.258}{0.021} & \err{0.501}{0.014} \\
        \bf\mbox{Qwen2.5-VL-72B-Instruct}                   & 73.4B     & \err{0.730}{0.017} & \err{0.378}{0.023} & \err{0.554}{0.015} \\
        \bf\mbox{QVQ-72B-Preview}                           & 73.4B     & \err{0.717}{0.022} & \err{0.139}{0.019} & \err{0.428}{0.017} \\
        \bf\mbox{llava-onevision-qwen2-72b-ov-chat}         & 73.2B     & \err{0.382}{0.021} & \err{0.185}{0.020} & \err{0.281}{0.014} \\
        \bf\mbox{gemma-3-27b-it}                            & 27.4B     & \err{0.746}{0.021} & \err{0.250}{0.021} & \err{0.499}{0.015} \\
        \bf\mbox{Aria}                                      & 25.3B     & \err{0.258}{0.022} & \err{0.074}{0.014} & \err{0.165}{0.013} \\
        \bf\mbox{MiniCPM-V-2\_6}                            & 8.1B      & \err{0.091}{0.015} & \err{0.018}{0.006} & \err{0.054}{0.009} \\
        \bf\mbox{Qwen2.5-VL-3B-Instruct}                    & 3.75B     & \err{0.023}{0.007} & \err{0.003}{0.002} & \err{0.013}{0.004} \\
        \bottomrule
    \end{tabularx}
    \label{tab:OCR results}
\end{table}

\subsection{Complete Results on Extended Data}\label{experiment_extension_appendix}
In this section, we present additional results for evaluations on the extended data.
As shown in Table~\ref{tab:extension_results_full}, LLMs generally perform better across these three categories than on the English puzzles; however, even for puzzles from CommonSenseQA, performance remains far from perfect.
In contrast, LVLMs do not exhibit a significant performance improvement over English puzzles. 
Based on our observation of a strong positive correlation between grid-parsing and puzzle-solving performance (See Figure~\ref{fig: OCR_performance_correlation} for more details), we hypothesize that the reasoning capabilities of LVLMs are constrained by their visual processing abilities.
\begin{table}[htbp]
    \centering
    \footnotesize
        \caption{WCR and ICR for Chinese, English Simple, and CommonSenseQA puzzle sets.}
    \setlength{\tabcolsep}{2pt} 
    \renewcommand{\arraystretch}{1.3} 
    \resizebox{\linewidth}{!}{
    \begin{tabularx}{\textwidth}{@{}Xcc|cc|cc}
        \toprule
        \multirow{2}{*}{\textbf{Models}} 
         & \multicolumn{2}{c}{\textbf{Chinese}} 
         & \multicolumn{2}{c}{\textbf{English Simple}} 
         & \multicolumn{2}{c}{\textbf{CommonSenseQA}} \\
        \cmidrule(lr){2-3} \cmidrule(lr){4-5} \cmidrule(lr){6-7}
         & \textbf{WCR} & \textbf{ICR} 
         & \textbf{WCR} & \textbf{ICR} 
         & \textbf{WCR} & \textbf{ICR} \\
        \midrule[\heavyrulewidth]
        \addlinespace[0.5em]
        \multicolumn{7}{c}{\cellcolor{customBlue}\textit{\textbf{Proprietary LVLMs}}} \\
        \midrule
        \bf\mbox{GPT-4o-2024-11-20} 
         & \err{0.366}{0.018} & \err{0.170}{0.017} 
         & \err{0.335}{0.015} & \err{0.227}{0.017} 
         & \err{0.392}{0.026} & \err{0.247}{0.026} \\
        \bf\mbox{Claude 3.7 sonnet} 
         & \err{0.339}{0.023} & \err{0.267}{0.018} 
         & \err{0.408}{0.018} & \err{0.288}{0.018} 
         & \err{0.540}{0.022} & \err{0.386}{0.028}  \\
        \bf\mbox{Gemini 2.0 Flash}  
         & \err{0.208}{0.031} & \err{0.233}{0.018} 
         & \err{0.229}{0.014} & \err{0.216}{0.013} 
         & \err{0.327}{0.021} & \err{0.216}{0.018} \\
        
        \midrule[\heavyrulewidth]
        \addlinespace[0.5em]
        \multicolumn{7}{c}{\cellcolor{customBlue}\textit{\textbf{Open-Weight LVLMs}}} \\
        \midrule
        \bf\mbox{Pixtral-Large-Instruct-2411} 
         & \err{0.252}{0.015} & \err{0.101}{0.011} 
         & \err{0.216}{0.016} & \err{0.187}{0.015} 
         & \err{0.439}{0.023} & \err{0.270}{0.023} \\
        \bf\mbox{Qwen2.5-VL-72B-Instruct} 
         & \err{0.391}{0.025} & \err{0.282}{0.018} 
         & \err{0.239}{0.016} & \err{0.183}{0.015}  
         & \err{0.418}{0.022} & \err{0.252}{0.023}  \\
        
        \midrule[\heavyrulewidth]
        \addlinespace[0.5em]
        \multicolumn{7}{c}{\cellcolor{customGreen}\textit{\textbf{Proprietary LLMs}}} \\
        \midrule
        \bf\mbox{GPT-4o-2024-11-20} 
         & \err{0.593}{0.019} & \err{0.448}{0.021} 
         & \err{0.438}{0.020} & \err{0.306}{0.022} 
         & \err{0.524}{0.024} & \err{0.326}{0.029} \\
        \bf\mbox{Claude 3.7 sonnet} 
         & \err{0.478}{0.019} & \err{0.470}{0.019}  
         & \err{0.539}{0.021} & \err{0.487}{0.021} 
         & \err{0.583}{0.024} & \err{0.518}{0.025}  \\
        \bf\mbox{o3-mini-high \faLightbulb[regular]} 
         & \err{0.774}{0.016} & \err{0.953}{0.014} 
         & \err{0.782}{0.021} & \err{0.946}{0.012} 
         & \err{0.812}{0.027} & \err{0.971}{0.011}  \\
        
        \midrule[\heavyrulewidth]
        \addlinespace[0.5em]
        \multicolumn{7}{c}{\cellcolor{customGreen}\textit{\textbf{Open-Weight LLMs}}} \\
        \midrule
        \bf\mbox{DeepSeek-R1 \faLightbulb[regular]} 
         & \err{0.907}{0.016} & \err{0.898}{0.018} 
         & \err{0.759}{0.017} & \err{0.787}{0.020}
         & \err{0.752}{0.031} & \err{0.829}{0.026}  \\
        \bf\mbox{QwQ-32B \faLightbulb[regular]} 
         & \err{0.701}{0.020} & \err{0.654}{0.022} 
         & \err{0.647}{0.021} & \err{0.734}{0.020} 
         & \err{0.699}{0.026} & \err{0.766}{0.027} \\
        \bottomrule
    \end{tabularx}}
    \label{tab:extension_results_full}
\end{table}

\subsection{More on Test-Time Scaling}
In Section~\ref{experiment:TTS}, we use a bar chart to demonstrate the performance differences associated with three distinct reasoning efforts for o3-mini on \texttt{7x7} English puzzles.
Here, we provide a table detailing the specific numerical values for each metric as a supplement to Figure~\ref{fig:TTS}.
\begin{table}[htbp]
    \centering
    \footnotesize
        \caption{o3-mini performance on \texttt{7x7} English puzzles across three distinct reasoning efforts.}
    \begin{tabularx}{\linewidth}{@{}Xc|c|c|c}
        \toprule
        \textbf{Reasoning Effort} & \textbf{WCR} & \textbf{LCR} & \textbf{ICR} & \textbf{Avg. Tokens}\\
        \midrule
        \bf\mbox{High}                      & \err{0.587}{0.023} & \err{0.684}{0.021} & \err{0.891}{0.018} & 49865 \\
        \bf\mbox{Medium}                    & \err{0.534}{0.022} & \err{0.634}{0.022} & \err{0.777}{0.025} & 23723 \\
        \bf\mbox{Low}                       & \err{0.295}{0.018} & \err{0.392}{0.019} & \err{0.363}{0.026} & 4770 \\
        \bottomrule
    \end{tabularx}
    \label{tab:small TTS results}
\end{table}

\subsection{Token Usage}
In this section, we report the token usage for all evaluated models on both the \texttt{7x7} and \texttt{14x14} English puzzles.
Notably, we include non-reasoning models in this analysis, defining token usage as the total number of completion tokens.
For reasoning models, token usage is calculated as the sum of reasoning tokens and response tokens.
As shown in Figure~\ref{fig:token_usage_combined}, token usage increases with grid size across all models, with reasoning models generating more.
\begin{figure}[htbp]
    \centering
    \begin{subfigure}{\textwidth}
        \centering
        \includesvg[width=\textwidth]{graphs/token_usage_comparison.svg}
        \caption{Token Usage of LVLMs.}
        \label{fig:Token_Usage}
    \end{subfigure}
    \vspace{1em} 
    \begin{subfigure}{\textwidth}
        \centering
        \includesvg[width=\textwidth]{graphs/token_usage_comparison_text.svg}
        \caption{Token Usage of LLMs.}
        \label{fig:Text_Token_Usage}
    \end{subfigure}
    \caption{Token usage on \texttt{7x7} and \texttt{14x14} English puzzles. Reasoning models are in red.}
    \label{fig:token_usage_combined}
\end{figure}

\subsection{Ablations: LVLM Input Format and Modality Effects}\label{appendix: LVLM format}
To isolate modality-specific effects, we explore feeding LLMs the OCR-extracted output from grid images.
Our experiments show that while both OCR models and LVLMs can accurately extract clue text from images, they struggle to follow instructions and to construct a spatially coherent representation of the empty grid.
For example, when applying \href{https://huggingface.co/stepfun-ai/GOT-OCR-2.0-hf}{stepfun-ai/GOT-OCR-2.0-hf} to an image containing both the grid and clues, it generates \textit{“Across:\textbackslash n2. Canon product, for short\textbackslash n4. Defense aid\textbackslash n6. Certain fraud protector, for short\textbackslash n7. Abbr. before a founding date\textbackslash n10. Philosopher's study\textbackslash n12. May honoree\textbackslash n13. Fraternity letter\textbackslash nDown:\textbackslash n1. Like white panthers\textbackslash n2. [Not my mistake]\textbackslash n3. Beta dog's view\textbackslash n5. Gridiron abbr.\textbackslash n8. One of the muskrats in the 1976 hit "Muskrat Love"\textbackslash n9. Slow-witted\textbackslash n11. Going rate?: Abbr.\textbackslash n”}.
While the clue text is correctly extracted, the output lacks any representation of the grid structure, rendering it unsuitable as input for LLMs due to the absence of critical structural constraints.
On the other hand, the clue content extracted by OCR models and LVLMs is nearly identical to what we explicitly feed to LLMs in the text input setting. In fact, we augment the clue text with explicit positional hints to compensate for the lack of visual spatial information---details that LVLMs might infer from the image but LLMs cannot derive from plain text alone.

In our main evaluations, both the clues and the grid are embedded within a single image as inputs for LVLMs. 
In this section, we investigate the effect of isolating the clues and providing them to the LVLMs as separate text inputs, while the image contains only an empty grid.
The prompt used for this experimental setting is shown in Figure~\ref{Image CoT Grid only} in Appendix~\ref{implementation_details_appendix}.

The results, shown in Figure~\ref{fig:prompt_ablations} and measured by WCR on \texttt{7x7} English puzzles, reveal no significant performance differences between the two input formats. This indicates that the input format is not the primary cause of the suboptimal performance observed for LVLMs on CrossWordBench, suggesting that LVLMs are robust to slight variations in input format.
\begin{figure}[htbp]
    \centering
    \vspace{-1.0em}
    \includesvg[width=1.0\textwidth]{graphs/grid_vs_regular_cot_comparison.svg}
    \vspace{-2.5em}
    \caption{WCR difference on two inputs formats for LVLMs.}
    \label{fig:prompt_ablations}
\end{figure}

\subsection{More on Crossing Letter Count}\label{appendix_crossing_letter_constraints}
\begin{wraptable}{r}{0.5\textwidth}
\centering
\footnotesize
\vspace{-4em}
\caption{Average WCR results by crossing letter count on \texttt{7x7} and \texttt{14x14} English puzzles.}
\begin{tabular}{@{}lccc@{}}
\toprule
\textbf{Grid} & \textbf{Crossing Letter} & \textbf{Reason.} & \textbf{Non-Reason.} \\
\midrule
7x7    & Low   & 0.4539 & 0.2922 \\
7x7    & Med   & 0.5308 & 0.3230 \\
7x7    & High  & 0.6183 & 0.3259 \\
14x14  & Low   & 0.4115 & 0.3469 \\
14x14  & Med   & 0.4186 & 0.3514 \\
14x14  & High  & 0.2306 & 0.1438 \\
\bottomrule
\end{tabular}
\vspace{-2.0em}
\label{tab:appendix_crossing_letter_constraints}
\end{wraptable}
Table~\ref{tab:appendix_crossing_letter_constraints} presents the results shown in Figure~\ref{fig:intersection_accuracy}, along with additional results on \texttt{14x14} English puzzles.
On these larger puzzles, even reasoning LLMs struggle to effectively leverage crossing-letter constraints.
We hypothesize that this is due to the increased complexity of larger puzzles, which remains challenging for reasoning models.

\subsection{More on Few-Shot Prompting and Prefill Ratio Control}\label{appendix: few_shot}
In our main evaluation, we treat each puzzle as a whole rather than as a set of independent question–answer (QA) pairs, in order to preserve the structural constraints inherent to crossword solving.
This setup is not well-suited to few-shot prompting, as providing a full puzzle solution as a demonstration example would be lengthy and potentially distracting for the model.
Nevertheless, we experiment with one-shot prompting on \texttt{7x7} English puzzles, using a completed example generated by DeepSeek-R1 on the English\_Simple subset as the demonstration.
As shown in Figure~\ref{tab:few-shot}, the performance difference is not significant.

As discussed in Section~\ref{grid_format_ablation}, our controllable generation framework supports difficulty adjustment via the prefill ratio.
In Table~\ref{tab:wcr-prefill}, we compare performance on \texttt{7x7} English puzzles under two prefill conditions: 0\% and 50\%. To ensure a fair comparison, we prevent any word from being fully revealed.
As expected, a higher prefill ratio leads to improved performance, as more letters are made available to the model.
These results suggest that prefill ratio is a more effective method for puzzle difficulty control than few-shot prompting.
\begin{table}[htbp]
    \centering
    \caption{WCR under different prompting and prefill ratio settings on \texttt{7x7} English puzzles.}
    \begin{subtable}[c]{0.48\textwidth}
        \centering
        \caption{WCR with 0-shot and 1-shot prompting.}
        \begin{tabular}{@{}lcc@{}}
            \toprule
            \textbf{Model} & \textbf{0-Shot} & \textbf{1-Shot} \\
            \midrule
            GPT-4o (Text) & 0.410 & 0.394 \\
            Claude-Sonnet (Text) & 0.482 & 0.519 \\
            Gemini-Flash (Text) & 0.301 & 0.313 \\
            Llama-3.3-70B (Text) & 0.303 & 0.269 \\
            DeepSeek-V3 (Text) & 0.303 & 0.307 \\
            \bottomrule
        \end{tabular}\label{tab:few-shot}
    \end{subtable}
    \hfill
    \begin{subtable}[c]{0.48\textwidth}
        \centering
        \caption{WCR with 0\% and 50\% prefill.}
        \begin{tabular}{@{}lcc@{}}
            \toprule
            \textbf{Model} & \textbf{0\%} & \textbf{50\%} \\
            \midrule
            GPT-4o (Text) & 0.410 & 0.483 \\
            GPT-4o (Img) & 0.348 & 0.383 \\
            Claude-Sonnet (Text) & 0.482 & 0.604 \\
            Claude-Sonnet (Img) & 0.479 & 0.568 \\
            Gemini-Flash (Text) & 0.301 & 0.362 \\
            Gemini-Flash (Img) & 0.277 & 0.346 \\
            o3-mini (Text) & 0.587 & 0.864 \\
            \bottomrule
        \end{tabular}\label{tab:wcr-prefill}
    \end{subtable}
\end{table}

\section{Implementation Details}\label{implementation_details_appendix}
\subsection{Evaluation Metrics}
Here, we provide a more complete and formal description of the three metrics used.
\begin{itemize}[leftmargin=*,itemsep=0pt]
\item \textbf{Word Coverage Rate (WCR):}
WCR measures word-level accuracy by calculating the percentage of correctly solved words in the crossword puzzle, defined as:
\begin{equation}
    \mathrm{WCR}=\frac{1}{\left|\mathcal{W}_A\right|+\left|\mathcal{W}_D\right|}\left(\sum_{w \in \mathcal{W}_A} \mathbb{1}\left\{r_w=m_w\right\}+\sum_{w \in \mathcal{W}_D} \mathbb{1}\left\{r_w=m_w\right\}\right) .
\end{equation}\label{WCR}where $\mathcal{W}_A$ and $\mathcal{W}_D$ denote the set of across and down words, respectively. For each word $w$, $r_w$ represents the reference answer, while $m_w$ represents the model answer.

\item \textbf{Letter Coverage Rate (LCR):}
LCR evaluates letter-level accuracy, providing partial credit for correct letter placements. For each word $w$, Let:
\[
C_w=\sum_{j=1}^{\min \left(\left|r_w\right|,\left|m_w\right|\right)} 
\mathbb{1}\left\{r_w[j]=m_w[j]\right\}
\textbf{ and }
L_w=\max \left(\left|r_w\right|,\left|m_w\right|\right)
\]where $C_w$ counts the correctly matched letters and $L_w$ is the total number of positions considered, the overall letter accuracy is defined as:
\begin{equation}
\centering
\begin{aligned}
\mathrm{LCR}&=\frac{\sum_{w \in \mathcal{W}} C_w}{\sum_{w \in \mathcal{W}} L_w}
\end{aligned}
\end{equation}\label{LCR}

\item \textbf{Intersection Consistency Rate (ICR)}:
the internal consistency of the model's answers at intersections where across and down words overlap, defined as:
\vspace*{-0.5\baselineskip}
\begin{equation}
\mathrm{ICR}=\frac{1}{|\mathcal{I}|} \sum_{(a, d, j, k) \in \mathcal{I}} \mathbb{1}\{a[j]=d[k]\}
\end{equation}\label{ICR_appendix}where $\mathcal{I}$ denotes the set of all intersections, where each tuple $(a, d, j, k)$ indicates the $j$th letter of the across word $a$ overlaps with the $k$th letter of the down word $d$.
This metric reflects whether models correctly adhere to the grid structural constraints of a puzzle.
\end{itemize}

\subsection{Parsing Details}\label{appendix parsing}
Answers extracted from model responses are converted into JSON format by leveraging the structured output capabilities of o3-mini and dynamic Pydantic models that adhere to the reference answer structure. Algorithm~\ref{alg:create_dynamic_puzzle_model} provides the pseudocode for creating these models.
\begin{algorithm}[H]
\footnotesize
\caption{Create Dynamic Pydantic Model}
\label{alg:create_dynamic_puzzle_model}
\begin{algorithmic}[1]
\Function{Create\_Dynamic\_Pydantic\_Model}{\texttt{reference\_answers}}
    \State \texttt{fields} $\gets \{\}$ \Comment{Initialize an empty dictionary}
    \State \texttt{pattern} $\gets$ \verb/^(across|down)\s+\d+\$/
    \ForAll{$answer \in \texttt{reference\_answers}$}
        \State \texttt{key} $\gets answer[\texttt{"direction"}]$
        \If{\texttt{key} does not match \texttt{pattern}}
            \State \textbf{raise} ValueError(\texttt{"Reference key does not match expected format"})
        \EndIf
        \State \texttt{fields}[\texttt{key}] $\gets$ (\texttt{Optional[ClueAnswer]}, Field with description ``Answer for clue at \texttt{key}'')
    \EndFor
    \State \texttt{Dynamic\_Pydantic\_Model} $\gets$ create\_model(\texttt{"Dynamic\_Pydantic\_Model"}, **\texttt{fields})
    \State \Return \texttt{Dynamic\_Pydantic\_Model}
\EndFunction
\end{algorithmic}
\end{algorithm}

\subsection{Generation Configuration}\label{appendix_generation_details}
Table~\ref{tab:generation_config} lists all the models evaluated along with their corresponding generation configs.
\begin{table}[htbp]
    \footnotesize
    \centering
    \renewcommand{\arraystretch}{1.2}
    \setlength{\tabcolsep}{2pt}
    \begin{tabularx}{\textwidth}{Xccc}
        \toprule
        \textbf{Model Name} & \textbf{Max Tokens (7×7)} & \textbf{Max Tokens (14×14)} & \textbf{Temperature} \\
        \midrule
        Claude 3.7 Sonnet~\citep{anthropic_claude_2024}                                                             & 8192      & 8192      & 0.0 \\
        Claude 3.7 Sonnet (Thinking)~\citep{anthropic_claude_2024}                                                  & 64000     & 64000     & 1.0 \\
        GPT-4o-2024-11-20~\citep{hurst2024gpt}                                                                      & 16384     & 16384     & 0.0 \\
        Gemini 2.0 Pro Exp~\citep{gemini_2}                                                                         & 20480     & 20480     & 0.0 \\
        Gemini 2.0 Flash~\citep{gemini_2}                                                                           & 20480     & 20480     & 0.0 \\
        o3-mini-high~\citep{o3_mini}                                                                                & 100000    & 100000    & 0.6 \\
        Pixtral-Large-Instruct-2411~\citep{agrawal2024pixtral}                                                      & 20480     & 20480     & 0.0 \\
        NVLM-D-72B~\citep{nvlm2024}                                                                                 & 20480     & 20480     & 0.0 \\
        InternVL2 5-78B-MPO~\citep{chen2024internvl}                                                                & 20480     & 20480     & 0.0 \\
        Qwen2.5-VL-72B-Instruct~\citep{bai2025qwen2}                                                                & 20480     & 20480     & 0.0 \\
        QVQ-72B-Preview~\citep{qvq-72b-preview}                                                                     & 100000    & 100000    & 0.0 \\
        llava-onevision-72b-ov-chat~\citep{li2024llava}                                                             & 20480     & 20480     & 0.0 \\
        gemma-3-27b-it~\citep{gemma_2025}                                                                           & 8192      & 8192     & 0.0 \\
        Aria~\citep{aria}                                                                                           & 20480     & 20480     & 0.0 \\
        MiniCPM-V-2\_6~\citep{yao2024minicpm}                                                                       & 20480     & 20480     & 0.0 \\
        Qwen2.5-VL-3B-Instruct~\citep{bai2025qwen2}                                                                 & 20480     & 20480     & 0.0 \\
        Llama-3.1-405B-Instruct~\citep{dubey2024llama}                                                              & 100000    & 100000    & 0.0 \\
        DeepSeek-R1~\citep{guo2025deepseek}                                                                         & 100000    & 100000    & 0.6 \\
        DeepSeek-V3~\citep{liu2024deepseek}                                                                         & 100000    & 100000    & 0.0 \\
        R1-Distill-Llama-70B~\citep{guo2025deepseek}                                                                & 100000    & 100000    & 0.6 \\
        Llama-3.3-70B-Instruct~\citep{dubey2024llama}                                                               & 20480     & 20480     & 0.0 \\
        QwQ-32B~\citep{qwq32b}                                                                     & 100000    & 100000    & 0.6 \\
        Phi-4~\citep{abdin2024phi}                                                                       & 10000     & 10000     & 0.0 \\
        \bottomrule
    \end{tabularx}
    \caption{Generation Configurations for CrossWordBench.}
    \label{tab:generation_config}
\end{table}


\FloatBarrier
\subsection{Prompts}\label{appendix prompts}
Figures~\ref{Image CoT}, \ref{Image CoT Grid only}, \ref{Text CoT}, \ref{Interactive Prompt}, \ref{Interactive Follow-up Prompt}, \ref{OCR Prompt}, \ref{Answer Parsing Prompt}, and \ref{Claude System Prompt} present all the prompts employed in this study—comprising image prompts for LVLMs and text prompts for LLMs. We observe that Claude 3.7 Sonnet sometimes produces partial outputs and requests confirmation to continue. To mitigate this issue, we incorporate an additional system prompt (see Figure~\ref{Claude System Prompt} for the system prompt); note that this modification applies only to the Claude 3.7 Sonnet.
\begin{figure*}[htbp]
\scriptsize
\centering
\begin{tcolorbox}[colback=white, colframe=customBlue, width=1.0\textwidth, arc=3mm, boxrule=0.5mm, title=Image Zero-Shot CoT Prompt]
\begin{Verbatim}[breaklines=true, breakanywhere=true, formatcom=\bfseries]
You are given a crossword puzzle image containing clues and a grid. Your task is to solve the puzzle accurately, ensuring that all answers fit both the given clues and the grid structure, including intersecting words.

For each clue, provide a step-by-step explanation:
- Identify the clue by its EXACT NUMBER AND DIRECTION as shown in the image. The numbers may not be sequential (e.g., Across clues might be numbered 1, 4, 7, and 9, while Down clues might be 2, 3, 5, 6, and 8).
- Determine word length from available grid spaces.
- Check for any pre-filled letters from intersecting words that have already been solved and explain how they constrain possible answers.
- Analyze the clue (definition, wordplay, cryptic hint).
- Explain your reasoning process.
- Confirm alignment with crossing letters.

Solving tips:
- Answers must be a single word with no spaces (combine phrases if needed).
- Abbreviations in clues typically indicate abbreviated answers.
- Match the clue's tense, singular/plural form, and part of speech.
- Look for wordplay signals, such as question marks (?) for puns or cryptic hints.
- Down words are filled from top to bottom, Across words from left to right.
- Always confirm that intersecting words remain valid after placing each answer.

Present your final solution as:
Across:
[Number as shown in image]: [Answer]

Down:
[Number as shown in image]: [Answer]

IMPORTANT:
- DO NOT list clues in sequential numerical order. You MUST match the exact numbering pattern from the image.
- DO NOT ask for confirmation or stop midway. Always provide a complete solution for all clues.
\end{Verbatim}

\end{tcolorbox}
\caption{Image Zero-Shot CoT Prompt.}
\label{Image CoT}
\end{figure*}

\begin{figure*}[htbp]
\scriptsize
\centering
\begin{tcolorbox}[colback=white, colframe=customBlue, width=1.0\textwidth, arc=3mm, boxrule=0.5mm, title=Image Zero-Shot CoT with Grid Only Prompt]
\begin{Verbatim}[breaklines=true, breakanywhere=true, formatcom=\bfseries]
You are given a crossword puzzle image containing only the grid, while the clues are provided separately in text form. Your task is to solve the puzzle accurately, ensuring that all answers fit both the given clues and the grid structure, including intersecting words.

Clues:
Each clue contains:
- Clue Direction and Number (e.g., "Across 1", "Down 2").
- Start Position (row, column) for the first letter of the answer.
- The actual clue text.

<clues>

For each clue, provide a step-by-step explanation:
- Identify the clue by its EXACT NUMBER AND DIRECTION as shown in the clue description. The numbers may not be sequential (e.g., Across clues might be numbered 1, 4, 7, and 9, while Down clues might be 2, 3, 5, 6, and 8).
- Determine word length from available grid spaces.
- Check for any pre-filled letters from intersecting words that have already been solved and explain how they constrain possible answers.
- Analyze the clue (definition, wordplay, cryptic hint).
- Explain your reasoning process.
- Confirm alignment with crossing letters.

Solving tips:
- Answers must be a single word with no spaces (combine phrases if needed).
- Abbreviations in clues typically indicate abbreviated answers.
- Match the clue's tense, singular/plural form, and part of speech.
- Look for wordplay signals, such as question marks (?) for puns or cryptic hints.
- Down words are filled from top to bottom, Across words from left to right.
- Always confirm that intersecting words remain valid after placing each answer.

Present your final solution as:
Across:
[Number as shown in clues]: [Answer]

Down:
[Number as shown in clues]: [Answer]

IMPORTANT:
- DO NOT list clues in sequential numerical order. You MUST match the exact numbering pattern from the given clues.
- DO NOT ask for confirmation or stop midway. Always provide a complete solution for all clues.
\end{Verbatim}

\end{tcolorbox}
\caption{Image Zero-Shot CoT with Grid Only Prompt.}
\label{Image CoT Grid only}
\end{figure*}

\begin{figure*}[htbp]
\scriptsize
\centering
\begin{tcolorbox}[colback=white, colframe=customBlue, width=1.0\textwidth, arc=3mm, boxrule=0.5mm, title=Text Zero-Shot CoT Prompt]
\begin{Verbatim}[breaklines=true, breakanywhere=true, formatcom=\bfseries]
You are given a crossword puzzle grid and a set of clues. Your task is to solve the puzzle accurately, ensuring that all answers fit both the given clues and the grid structure, including intersecting words.

Grid Representation:
The crossword grid is represented as a 2D array where:
- `1` represents a black (blocked) cell
- `0` represents an empty (unfilled) cell

<grid>

Clues:
Each clue contains:
- Clue Direction and Number (e.g., "Across 1", "Down 2").
- Start Position (row, column) for the first letter of the answer.
- The actual clue text.

<clues>

For each clue, provide a step-by-step explanation:
- Identify the clue by its EXACT NUMBER AND DIRECTION as shown in the clue description. The numbers may not be sequential (e.g., Across clues might be numbered 1, 4, 7, and 9, while Down clues might be 2, 3, 5, 6, and 8).
- Determine word length from available grid spaces.
- Check for any pre-filled letters from intersecting words that have already been solved and explain how they constrain possible answers.
- Analyze the clue (definition, wordplay, cryptic hint).
- Explain your reasoning process.
- Confirm alignment with crossing letters.

Solving tips:
- Answers must be a single word with no spaces (combine phrases if needed).
- Abbreviations in clues typically indicate abbreviated answers.
- Match the clue's tense, singular/plural form, and part of speech.
- Look for wordplay signals, such as question marks (?) for puns or cryptic hints.
- Down words are filled from top to bottom, Across words from left to right.
- Always confirm that intersecting words remain valid after placing each answer.

Present your final solution as:
Across:
[Number as shown in clues]: [Answer]

Down:
[Number as shown in clues]: [Answer]

IMPORTANT:
- DO NOT list clues in sequential numerical order. You MUST match the exact numbering pattern from the given clues.
- DO NOT ask for confirmation or stop midway. Always provide a complete solution for all clues.
\end{Verbatim}

\end{tcolorbox}
\caption{Text Zero-Shot CoT Prompt.}
\label{Text CoT}
\end{figure*}

\begin{figure*}[htbp]
\scriptsize
\centering
\begin{tcolorbox}[colback=white, colframe=customBlue, width=1.0\textwidth, arc=3mm, boxrule=0.5mm, title=Interactive Mode Prompt]
\begin{Verbatim}[breaklines=true, breakanywhere=true, formatcom=\bfseries]
You are given a crossword puzzle image containing clues and a grid. Your task is to solve the puzzle accurately, ensuring that all answers fit both the given clues and the grid structure, including intersecting words.

Pick ONE clue, provide a step-by-step explanation:
- Identify the clue by its EXACT NUMBER AND DIRECTION as shown in the image. The numbers may not be sequential (e.g., Across clues might be numbered 1, 4, 7, and 9, while Down clues might be 2, 3, 5, 6, and 8).
- Determine word length from available grid spaces.
- Check for any pre-filled letters from intersecting words that have already been solved and explain how they constrain possible answers.
- Analyze the clue (definition, wordplay, cryptic hint).
- Explain your reasoning process.
- Confirm alignment with crossing letters.

Solving tips:
- Answers must be a single word with no spaces (combine phrases if needed).
- Abbreviations in clues typically indicate abbreviated answers.
- Match the clue's tense, singular/plural form, and part of speech.
- Look for wordplay signals, such as question marks (?) for puns or cryptic hints.
- Down words are filled from top to bottom, Across words from left to right.
- Always confirm that intersecting words remain valid after placing each answer.

Only solve ONE clue at a time and wait for confirmation before proceeding to the next round.
\end{Verbatim}

\end{tcolorbox}
\caption{Interactive Mode Prompt.}
\label{Interactive Prompt}
\end{figure*}

\begin{figure*}[htbp]
\scriptsize
\centering
\begin{tcolorbox}[colback=white, colframe=customBlue, width=1.0\textwidth, arc=3mm, boxrule=0.5mm, title=Interactive Mode Follow-up Prompt]
\begin{Verbatim}[breaklines=true, breakanywhere=true, formatcom=\bfseries]
For the following round:

Using the confirmed answers so far:

Pick another clue, provide a step-by-step explanation:
- Identify the clue by its EXACT NUMBER AND DIRECTION as shown in the image. The numbers may not be sequential (e.g., Across clues might be numbered 1, 4, 7, and 9, while Down clues might be 2, 3, 5, 6, and 8).
- Determine word length from available grid spaces.
- Check for any pre-filled letters from intersecting words that have already been solved and explain how they constrain possible answers.
- Analyze the clue (definition, wordplay, cryptic hint).
- Explain your reasoning process.
- Confirm alignment with crossing letters.

Solving tips:
- Prioritize clues that intersect with confirmed answers.
- Answers must be a single word with no spaces (combine phrases if needed).
- Abbreviations in clues typically indicate abbreviated answers.
- Match the clue's tense, singular/plural form, and part of speech.
- Look for wordplay signals, such as question marks (?) for puns or cryptic hints.
- Down words are filled from top to bottom, Across words from left to right.
- Always confirm that intersecting words remain valid after placing each answer.

Only solve ONE clue at a time and wait for confirmation before proceeding to the next round and do not repeat previously solved clues.
\end{Verbatim}

\end{tcolorbox}
\caption{Interactive Mode Follow-up Prompt.}
\label{Interactive Follow-up Prompt}
\end{figure*}

\begin{figure*}[htbp]
\scriptsize
\centering
\begin{tcolorbox}[colback=white, colframe=customBlue, width=1.0\textwidth, arc=3mm, boxrule=0.5mm, title=Grid-Parsing Prompt]
\begin{Verbatim}[breaklines=true, breakanywhere=true, formatcom=\bfseries]
Your task is to extract and match all words from a crossword puzzle grid with their respective clues. The image consists of two sections:
1. The Clue Section:
- Contains two lists: "Across" and "Down."
- Each clue is numbered and corresponds to a starting position in the grid.
2. The Grid Section:
- A crossword grid containing letters, empty cells, and numbered starting positions for words.
- Words extend either across (left to right) or down (top to bottom).

Step 1: Extract Clues and Grid Structure
- Identify all clues under the "Across" and "Down" sections, preserving their numbers.
- Identify all numbered cells in the grid.

Step 2: Extract Words from the Grid
For each numbered cell:
- If the word extends ACROSS:
    - Start at the numbered cell and read consecutive letters left to right until reaching an empty cell or grid boundary.
- If the word extends DOWN:
    - Start at the numbered cell and read consecutive letters top to bottom until reaching an empty cell or grid boundary.

Step 3: Match Words to Clues
- Match each numbered word in the grid to its corresponding clue in the Across or Down section.
- Ensure extracted words are correctly assigned to their respective clues.

Output Format:
ACROSS:
[Number as shown in image]: [Clue Text]
Extracted Word: [Word from Grid]

DOWN:
[Number as shown in image]: [Clue Text]
Extracted Word: [Word from Grid]

Ensure accuracy in matching words to their clues, and extract all words fully without omitting any.
\end{Verbatim}

\end{tcolorbox}
\caption{Grid-Parsing Prompt.}
\label{OCR Prompt}
\end{figure*}

\begin{figure*}[htbp]
\scriptsize
\centering
\begin{tcolorbox}[colback=white, colframe=customBlue, width=1.0\textwidth, arc=3mm, boxrule=0.5mm, title=Self-Reflection Prompt]
\begin{Verbatim}[breaklines=true, breakanywhere=true, formatcom=\bfseries]
Your previous solution contains incorrect answers. Take a step back, carefully re-examine your entries, and systematically verify each word to ensure complete consistency and correctness within the crossword puzzle.

Provide a step-by-step verification:
1. Cross-Check Letters: List every intersection explicitly, noting the letters where Across and Down clues meet.
2. Consistency Check: Verify that each intersection matches perfectly. Identify and highlight any conflicting letters immediately.
3. Clue Validation: Revisit each clue thoroughly, confirming that each answer fully aligns with its clue description and adheres strictly to length constraints.
4. Grid Integrity: Confirm that your corrected entries maintain the integrity of the puzzle grid, leaving no unresolved conflicts or empty cells.

After completing these steps, present your revised and verified solutions in the following format:
Across:
[Clue Number]: [Corrected Answer]

Down:
[Clue Number]: [Corrected Answer]

IMPORTANT:
- DO NOT list clues in sequential numerical order. You MUST match the exact numbering pattern.
- Do NOT restate previous incorrect answers. Provide only fully corrected solutions after reflection.
\end{Verbatim}

\end{tcolorbox}
\caption{Self-Reflection Prompt.}
\label{Self-Reflection Prompt}
\end{figure*}

\begin{figure*}[htbp]
\scriptsize
\centering
\begin{tcolorbox}[colback=white, colframe=customBlue, width=1.0\textwidth, arc=3mm, boxrule=0.5mm, title=Answer-Parsing System Prompt]
\begin{Verbatim}[breaklines=true, breakanywhere=true, formatcom=\bfseries]
You are a crossword puzzle answer extractor. Extract only valid answers from a text response containing crossword solutions.

Requirements:
- If an answer contains spaces or multiple words, combine them into a single word.
- Do not shift or reorder answers. For example, if the expected keys are "Down 12" and "Down 13" and only the answer for "Down 13" is provided in the text, then "Down 12" should be null and "Down 13" should contain the answer.
- Do not invent or infer answers not explicitly stated in the text

Output your response in the given structure.
\end{Verbatim}

\end{tcolorbox}
\caption{Answer-Parsing Prompt.}
\label{Answer Parsing Prompt}
\end{figure*}

\begin{figure*}[htbp]
\scriptsize
\centering
\begin{tcolorbox}[colback=white, colframe=customBlue, width=1.0\textwidth, arc=3mm, boxrule=0.5mm, title=Claude System Prompt]
\begin{Verbatim}[breaklines=true, breakanywhere=true, formatcom=\bfseries]
You are a helpful assistant who completes tasks fully without seeking confirmation. Your role is to deliver comprehensive responses in one go. Never ask if the user wants you to continue or show more - you must provide the complete response.
\end{Verbatim}

\end{tcolorbox}
\caption{Claude System Prompt.}
\label{Claude System Prompt}
\end{figure*}

\FloatBarrier
\subsection{Textual Grid Representations}\label{appendix grid}
\begin{figure}[h]
\centering

\begin{subfigure}[b]{0.45\textwidth}
\centering
\[
\begin{bmatrix}
0 & 0 & 0 & 1 & 0 & 0 & 0 \\
0 & 1 & 0 & 0 & 1 & 0 & 1 \\
0 & 0 & 0 & 1 & 0 & 0 & 0 \\
0 & 1 & 1 & 1 & 0 & 1 & 0 \\
0 & 0 & 0 & 0 & 0 & 1 & 0 \\
1 & 0 & 1 & 0 & 1 & 1 & 0 \\
1 & 0 & 1 & 0 & 0 & 0 & 1
\end{bmatrix}
\]
\caption{2D array format (ARC-style).}
\end{subfigure}
\hfill
\begin{subfigure}[b]{0.5\textwidth}
\centering
\[
\begin{array}{c|ccccccc}
 & 0 & 1 & 2 & 3 & 4 & 5 & 6 \\
\hline
0 & \cdot & \cdot & \cdot & - & \cdot & \cdot & \cdot \\
1 & \cdot & - & \cdot & \cdot & - & \cdot & - \\
2 & \cdot & \cdot & \cdot & - & \cdot & \cdot & \cdot \\
3 & \cdot & - & - & - & \cdot & - & \cdot \\
4 & \cdot & \cdot & \cdot & \cdot & \cdot & - & \cdot \\
5 & - & \cdot & - & \cdot & - & - & \cdot \\
6 & - & \cdot & - & \cdot & \cdot & \cdot & -
\end{array}
\]
\caption{Markdown-style grid with symbols.}
\end{subfigure}

\caption{Two text representations of the puzzle grid: array (left) and markdown (right).}
\label{fig:grid_representations}
\end{figure}

\end{document}